\pdfoutput=1
\documentclass[journal]{IEEEtran}

\usepackage[utf8]{inputenc}
\usepackage[T1]{fontenc}
\usepackage{algorithm, algpseudocode}
\algrenewcommand{\algorithmiccomment}[1]{\hfill{\color{hookersgreen}\(\triangleright\) #1}}
\usepackage{cite, enumitem, graphicx, pgfplots, tikz, stmaryrd}
\usepackage[caption=false,font=footnotesize]{subfig}
\pgfplotsset{compat=1.18}
\usetikzlibrary{positioning, fit, arrows.meta, calc, backgrounds}

\definecolor{frenchblue}{RGB}{0,114,187}
\usepackage{amsfonts, amsmath, amssymb, amsthm}
\usepackage{booktabs, tabularx, url}
\usepackage{placeins}
\usepackage{CJKutf8, dsfont}
\usepackage{bm}
\usepackage[normalem]{ulem}
\PassOptionsToPackage{hyphens}{url}
\usepackage[colorlinks=true, citecolor=blue, linkcolor=blue, urlcolor=cyan]{hyperref}
\newlist{todolist}{itemize}{2}
\setlist[todolist]{label=$\square$}

\DeclareMathSizes{8.5}{8}{6}{5}
\input{Ustmry.fd}
\DeclareFontShape{U}{stmry}{b}{n}{<->ssub*stmry/m/n}{}


\definecolor{azure}{rgb}{0.0, 0.5, 1.0}
\definecolor{frenchblue}{rgb}{0.0, 0.45, 0.73}
\definecolor{forestgreen(traditional)}{rgb}{0.0, 0.27, 0.13}
\definecolor{mygreen}{rgb}{0.09, 0.45, 0.27}
\definecolor{myblue}{rgb}{0.2383,0.5195,0.7734}
\definecolor{mygreen}{rgb}{0.6445,0.9297,0.0039}
\definecolor{darklavender}{rgb}{0.45, 0.31, 0.59}
\definecolor{americanrose}{rgb}{1.0, 0.01, 0.24}
\definecolor{pigblue}{rgb}{0.2, 0.2, 0.6}
\definecolor{blue(ryb)}{rgb}{0.01, 0.28, 1.0}
\definecolor{amethyst}{rgb}{0.6, 0.4, 0.8}
\definecolor{deepmagenta}{rgb}{0.8, 0.0, 0.8}
\definecolor{carminered}{rgb}{1.0, 0.0, 0.22}
\definecolor{iris}{rgb}{0.35, 0.31, 0.81}

\newtheorem{assumption}{\hspace{0pt}\bf Assumption}
\newtheorem{lemma}{\hspace{0pt}\bf Lemma}
\newtheorem{proposition}{\hspace{0pt}\bf Proposition}

\newtheorem{theorem}{\hspace{0pt}\bf Theorem}
\newtheorem{corollary}{\hspace{0pt}\bf Corollary}

\def \s       {\, \text{\normalfont s} }

\def\ccal0{{\ensuremath{\mathcal 0}}}
\def\bb0{{\ensuremath{\boldsymbol 0}}}
\definecolor{crimson}{rgb}{0.86, 0.08, 0.24}
\definecolor{scarlet}{rgb}{1.0, 0.13, 0.0}
\definecolor{hookersgreen}{rgb}{0.0, 0.44, 0.0}
\definecolor{cultramarine}{rgb}{0.25, 0.0, 1.0}
\definecolor{applegreen}{rgb}{0.55, 0.71, 0.0}
\definecolor{ruddy}{rgb}{1.0, 0.0, 0.16}

\begin{document}

\title{Zeroth-Order Riemannian Optimization on Fixed-Rank Manifolds for LLM Fine-Tuning}

\author{Xinjue~Wang,~\IEEEmembership{Member,~IEEE},
        Xiuheng~Wang,~\IEEEmembership{Member,~IEEE},
        Yejun~Zhang,
        Sergiy~A.~Vorobyov,~\IEEEmembership{Fellow,~IEEE},
        Esa~Ollila,~\IEEEmembership{Senior Member,~IEEE},
        and~Jie~Chen,~\IEEEmembership{Senior Member,~IEEE}%
\thanks{Xinjue Wang, Sergiy A. Vorobyov, and Esa Ollila are with the Department of Information and Communications Engineering, Aalto University, 02150 Espoo, Finland (e-mail: xjw@ieee.org; sergiy.vorobyov@aalto.fi; esa.ollila@aalto.fi). Xiuheng Wang is with the School of Future Technology, South China University of Technology, Guangzhou 510640, China (e-mail: dr.xiuheng.wang@gmail.com). Yejun Zhang is with the Department of Computer Science, Aalto University, 02150 Espoo, Finland (e-mail: yejun.zhang@aalto.fi). Jie Chen is with the School of Artificial Intelligence, Northwestern Polytechnical University, Xi’an 710072, China, and also with the Research and Development Institute, Northwestern Polytechnical University, Shenzhen 518063, China (e-mail: dr.jie.chen@ieee.org).}
}

\maketitle

\begin{abstract}
Zeroth-order (ZO) optimization finds descent directions from function values alone without backpropagation, making it attractive for memory-efficient fine-tuning of large language models (LLMs).
Existing low-rank ZO methods apply low-rank perturbations in the full weight space or perturb coupled adapter factors, yet full-space steps accumulate high-rank updates while factor perturbations depend on nonunique factor scaling.
In this paper, we formulate forward-only LLM adaptation directly on fixed-rank matrix manifolds with frozen pretrained weights to preserve compact low-rank updates without factor-scaling ambiguity.
Within this geometry, we construct a normalized tangent ZO estimator from two forward passes per step and prove its parameterization invariance and alignment in expectation with the Riemannian gradient.
Building on this estimator and truncated singular value decomposition retraction, we develop zeroth-order Riemannian gradient descent (ZO-RGD) and its scalar-smoothed variant (ZO-RGD-S), and establish convergence guarantees for ZO-RGD on nonconvex objectives.
Numerical experiments on natural language understanding demonstrate that the proposed methods achieve competitive accuracy against forward-only baselines at inference-level memory cost.
\end{abstract}

\begin{IEEEkeywords}
Zeroth-order optimization, Riemannian optimization, fixed-rank matrix manifold, large language models, low-rank adaptation, natural language understanding.
\end{IEEEkeywords}

\section{Introduction}
\label{sec:intro}

\IEEEPARstart{L}{arge} language models (LLMs) have become foundational across natural language understanding (NLU), reasoning, and generation tasks~\cite{brown2020language, achiam2023gpt}. Adapting these models under constrained graphics processing unit (GPU) memory remains a central challenge in speech and language processing. Parameter-efficient fine-tuning (PEFT) methods such as low-rank adaptation (LoRA)~\cite{hu2022lora} freeze the pretrained weights and train low-rank additive updates, substantially reducing the number of trainable parameters and optimizer states~\cite{liu2024dora,zhang2023adaptiveLoRA,hayou2024loraplus}. However, backpropagation stores intermediate layer activations from the forward pass to compute exact gradients. On long sequences or large batches, these stored activations dominate peak memory. To avoid storing activations, forward-only zeroth-order (ZO) methods such as MeZO~\cite{malladi2023fine} estimate gradients from function evaluations at two points~\cite{spall1992spsa,ghadimi2013stochastic,Duchi2015RGDTIT,Nesterov2017RGE}. Regenerating perturbations in place from random seeds keeps peak memory at the inference level.

\begin{figure*}[!t]
\centering
\begin{tikzpicture}[
    x=1cm, y=1cm, >=Latex,
    font=\fontsize{8.5}{10}\selectfont,
    every node/.style={align=center, inner sep=1.5pt},
    updatearrow/.style={-{Latex[length=1.8mm,width=1.3mm]}, draw=updateorange, line width=1.1pt, shorten >=2pt},
    probearrow/.style={-{Latex[length=2.0mm,width=1.5mm]}, draw=queryblue, line width=1.2pt, shorten >=2pt},
    declare function={
        hh(\X,\Z)=-0.08*(\X-2.8)^2-0.04*(\Z-2.0)^2;
        edgeX(\Z)=6.8+0.4*(\Z-0.1)/3.8;
        sx(\X,\Z)=\X+0.5*\Z;
        sy(\X,\Z)=hh(\X,\Z)+0.3*\Z;
        ax(\X,\Y,\Z)=\X+0.5*\Z;
        ay(\X,\Y,\Z)=\Y+0.3*\Z;
        plx(\s,\t)=ax(4.2+\s,-0.1632-0.224*\s+0.032*\t,1.6+\t);
        ply(\s,\t)=ay(4.2+\s,-0.1632-0.224*\s+0.032*\t,1.6+\t);
        lnx(\s)=plx(\s,0.3*\s);
        lny(\s)=ply(\s,0.3*\s);
    }
]
\definecolor{surfaceedge}{RGB}{52,126,104}
\definecolor{queryblue}{RGB}{36,105,180}
\definecolor{normalizedindigo}{RGB}{77,70,167}
\definecolor{updateorange}{RGB}{194,73,37}
\definecolor{railblue}{RGB}{243,246,250}
\path[use as bounding box] (0,0) rectangle (13.9,4.60);
\begin{scope}[shift={(0.10,1.95)},yscale=1.25]

\fill[surfaceedge!14]
    plot[domain=0.5:6.8, samples=40] ({sx(\x,0.1)},{sy(\x,0.1)}) --
    plot[domain=0.1:3.9, samples=20] ({sx(edgeX(\x),\x)},{sy(edgeX(\x),\x)}) --
    plot[domain=7.2:0.5, samples=40] ({sx(\x,3.9)},{sy(\x,3.9)}) --
    plot[domain=3.9:0.1, samples=20] ({sx(0.5,\x)},{sy(0.5,\x)}) -- cycle;
\foreach \X in {1.6,2.8,4.0,5.2,6.4}
    \draw[surfaceedge!28, line width=0.35pt]
        plot[domain=0.1:3.9, samples=20] ({sx(\X,\x)},{sy(\X,\x)});
\foreach \Z in {1.0,2.0,3.0}
    \draw[surfaceedge!28, line width=0.35pt]
        plot[domain=0.5:{edgeX(\Z)}, samples=40] ({sx(\x,\Z)},{sy(\x,\Z)});
\draw[surfaceedge, line width=1.0pt]
    plot[domain=0.5:6.8, samples=40] ({sx(\x,0.1)},{sy(\x,0.1)}) --
    plot[domain=0.1:3.9, samples=20] ({sx(edgeX(\x),\x)},{sy(edgeX(\x),\x)}) --
    plot[domain=7.2:0.5, samples=40] ({sx(\x,3.9)},{sy(\x,3.9)}) --
    plot[domain=3.9:0.1, samples=20] ({sx(0.5,\x)},{sy(0.5,\x)}) -- cycle;
\node[text=surfaceedge!85!black, anchor=north] at (4.85,-1.02) {Rank-$r$ manifold $\mathcal M_r$};

\fill[railblue, fill opacity=0.62, draw=black!35, line width=0.5pt]
    ({plx(-3.9,-1.3)},{ply(-3.9,-1.3)}) -- ({plx(2.5,-1.3)},{ply(2.5,-1.3)}) --
    ({plx(2.5,1.4)},{ply(2.5,1.4)}) -- ({plx(-3.9,1.4)},{ply(-3.9,1.4)}) -- cycle;
\node[text=black!65, anchor=south] at (2.80,1.66) {Tangent space $T_{\Delta\boldsymbol W_t}\mathcal M_r$};

\coordinate (cur)   at ({lnx(0)},{lny(0)});
\coordinate (normtip) at ({lnx(2.25)},{lny(2.25)});
\coordinate (negtip) at ({lnx(-2.25)},{lny(-2.25)});
\coordinate (foot)  at ({lnx(0.85)},{lny(0.85)});
\coordinate (trial) at ({lnx(-3.2)},{lny(-3.2)});
\coordinate (ztip)  at ({ax(5.312172,0.824971,1.817547)},{ay(5.312172,0.824971,1.817547)});
\coordinate (next)  at ({sx(1.2025455708,0.7210789319)},{sy(1.2025455708,0.7210789319)});

\draw[-{Latex[length=1.7mm,width=1.2mm]}, draw=queryblue!45, line width=0.75pt]
    (cur) -- (ztip);
\node[text=queryblue!85!black, anchor=south] at ($(ztip)+(0,0.12)$)
    {Gaussian matrix $\boldsymbol Z_t$};
\draw[queryblue!75, densely dashed, line width=0.75pt,
    -{Latex[length=1.5mm,width=1.1mm]}, shorten >=3pt] (ztip) -- (foot);
\node[text=queryblue!85!black, anchor=west] at ($(ztip)+(0.08,-0.39)$) {Orthogonal projection};
\draw[normalizedindigo, line width=1.15pt,
    -{Latex[length=2.0mm,width=1.45mm]}, shorten >=1pt] (cur) -- (normtip);
\draw[probearrow, shorten >=3pt] (cur) -- (foot);
\node[text=queryblue, anchor=north east] at ($(foot)+(-0.12,0.02)$)
    {$\widehat{\boldsymbol Z}_t$};
\node[text=normalizedindigo, anchor=north] at ($(normtip)+(-0.38,-0.07)$)
    {$\widetilde{\boldsymbol Z}_t$};

\draw[updatearrow] (cur) -- (trial)
    node[pos=0.53, sloped, above=1.5pt, text=updateorange, inner sep=0pt]
    {ZO tangent\\step $-\eta\widehat{\boldsymbol G}_t$};
\draw[updatearrow, dashed] (trial) -- (next);
\node[text=updateorange, anchor=east] at (1.27,0.25) {SVD\\retraction};
\node[anchor=north west, inner sep=1pt] at (1.62,-0.24) {$\Delta\boldsymbol W_{t+1}$};

\draw[normalizedindigo, line width=1.15pt, dash pattern=on 3pt off 3pt,
    -{Latex[length=2.0mm,width=1.45mm]}, shorten >=1pt, shorten <=3pt] (cur) -- (negtip);
\node[text=normalizedindigo, anchor=north] at ($(negtip)+(0.82,-0.13)$)
    {$-\widetilde{\boldsymbol Z}_t$};

\node[circle, fill=updateorange, inner sep=0pt, minimum size=2.5pt] at (trial) {};
\node[circle, fill=updateorange, inner sep=0pt, minimum size=2.5pt] at (next) {};
\fill[queryblue] (foot) circle (0.045);
\fill[black!85] (cur) circle (0.075);
\node[anchor=north east] at (4.92,0.20) {$\Delta\boldsymbol W_t$};

\end{scope}

\draw[black!15, line width=0.5pt] (9.65,0.20) -- (9.65,4.44);
\node[anchor=north west, text=normalizedindigo, font=\bfseries\fontsize{8.5}{10}\selectfont]
    at (9.94,4.43) {Normalized random matrix};
\node[anchor=north west] at (9.94,4.03)
    {$\widetilde{\boldsymbol Z}_t=\sqrt{mn}\,\widehat{\boldsymbol Z}_t/\|\widehat{\boldsymbol Z}_t\|_{\mathrm F}$};
\node[anchor=north west, text=queryblue, font=\bfseries\fontsize{8.5}{10}\selectfont]
    at (9.94,3.35) {Two forward evaluations};
\node[anchor=north west] at (9.94,2.92)
    {$F_\pm=F(\Delta\boldsymbol W_t\pm\epsilon\widetilde{\boldsymbol Z}_t;\xi_t)$};
\node[anchor=north west] at (9.94,2.28)
    {$\hat s_t=\dfrac{F_+-F_-}{2\epsilon}$};
\node[anchor=north west, text=updateorange, font=\bfseries\fontsize{8.5}{10}\selectfont]
    at (9.94,1.20) {Update and retract};
\node[anchor=north west] at (9.94,0.82)
    {$\widehat{\boldsymbol G}_t=\hat s_t\widetilde{\boldsymbol Z}_t$};
\node[anchor=north west] at (9.94,0.39)
    {$\Delta\boldsymbol W_{t+1}=\mathcal R_{\Delta\boldsymbol W_t}(-\eta\widehat{\boldsymbol G}_t)$};
\end{tikzpicture}
\par\vspace{2pt}
\caption{Overview of one ZO-RGD step on an $m\times n$ rank-$r$ update block $\Delta\boldsymbol W_t\in\mathcal M_r$ (Section~\ref{sec:method}).
Projecting a standard Gaussian matrix $\boldsymbol Z_t$ onto $T_{\Delta\boldsymbol W_t}\mathcal M_r$ gives $\widehat{\boldsymbol Z}_t=P_{T_{\Delta\boldsymbol W_t}\mathcal M_r}(\boldsymbol Z_t)$, and Frobenius normalization gives $\widetilde{\boldsymbol Z}_t=\sqrt{mn}\,\widehat{\boldsymbol Z}_t/\|\widehat{\boldsymbol Z}_t\|_{\mathrm F}$.
Two forward passes on minibatch $\xi_t$ with radius $\epsilon>0$ evaluate the losses $F_\pm=F(\Delta\boldsymbol W_t\pm\epsilon\widetilde{\boldsymbol Z}_t;\xi_t)$.
Their difference yields the directional derivative $\hat s_t$ and the ZO Riemannian gradient $\widehat{\boldsymbol G}_t=\hat s_t\widetilde{\boldsymbol Z}_t$.
Truncated-SVD retraction $\mathcal R_{\Delta\boldsymbol W_t}$ then maps the step $-\eta\widehat{\boldsymbol G}_t$ with learning rate $\eta>0$ to $\Delta\boldsymbol W_{t+1}\in\mathcal M_r$.}
\label{fig:method_overview}
\end{figure*}
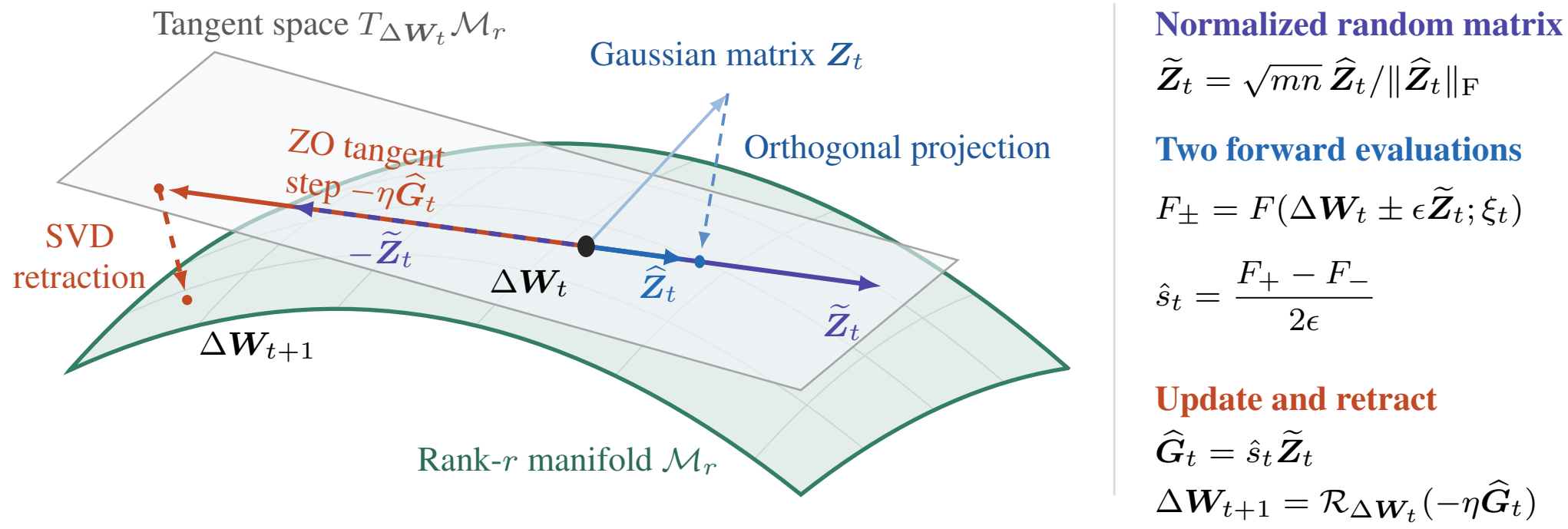

To combine inference-level training memory with low-rank structure, existing ZO methods follow two Euclidean routes. The first route updates the full weight matrices using low-rank perturbations to reduce estimator variance, as in LOZO~\cite{chen2025lozo} and random-subspace ZO (SubZero)~\cite{Yu2025ZORandomSubspace}. These methods periodically redraw their low-rank perturbation subspaces across iterations while adding every step directly to the full weight matrix. Consequently, the accumulated update becomes high-rank and cannot be stored as a compact low-rank adapter. To retain a compact adapter, the second route parameterizes each update through coupled matrix or tensor factors and perturbs them directly~\cite{malladi2023fine,jin2024dfolora,yang2024Adazeta}. However, any fixed-rank update admits infinitely many factorizations related by invertible rescaling. Such rescaling leaves the composite update unchanged, yet it alters isotropic perturbations and ZO update directions in factor space.

Riemannian optimization eliminates this factor ambiguity by treating fixed-rank matrices directly as a smooth manifold~\cite{absil2008optimization,vandereycken2013low,wei2016guarantees,schneider2015convergence}. In ZO optimization, however, standard Riemannian methods evaluate finite differences at retracted or exponential-map points on the manifold~\cite{li2022stochasticZO,li2023zorasa,he2024riemacc,ma2026structure}. On fixed-rank manifolds, retracting symmetric query points requires two extra singular value decomposition (SVD) operations per block at every step. For LLM adaptation, RoZO~\cite{rozo2026} avoids query-point retractions by perturbing the two low-rank factors with isotropic Gaussian noise to form ambient tangent directions. However, the resulting estimator weights tangent directions by the relative scaling of the two factors and the singular values of the update, so two factorizations of the same update matrix yield different search directions. A tangent ZO estimator without query-point retractions or factorization dependence is thus needed. Its convergence on fixed-rank manifolds for nonconvex objectives is also an open problem.

In this paper, we propose zeroth-order Riemannian optimization methods to optimize each additive LLM update block directly on a fixed-rank matrix manifold, as illustrated in Fig.~\ref{fig:method_overview}. At each step, we project a standard Gaussian perturbation orthogonally onto the local tangent space. Since this projection depends only on the current row and column subspaces, it removes factor-scaling ambiguity. Low-rank projection shrinks the perturbation norm, so we normalize each tangent direction to match the ambient norm. We then evaluate the loss with two forward passes along the ambient tangent line and retract the updated block onto the manifold.

\vspace{-.2cm}
\subsection{Related Work}
\label{subsec:related_work}

First-order PEFT adapts pretrained speech and language models by updating lightweight modules while keeping the backbone weights frozen~\cite{hu2022lora,li2026unifiedpeft}. Recent architectures extend low-rank adaptation through unified attention and feed-forward tuning~\cite{qi2025memorytuning}, weight-decomposed updates~\cite{liu2024dora,zhao2025weightdecomplora}, and singular-value tuning~\cite{vandereeckt2026svtuning}. For multi-task and multilingual transfer, mixture-of-experts and subspace-decoupled LoRA variants route inputs across shared and task-specific factors~\cite{lin2026hybridmoe,li2026mixmoe,mei2026zipperlora}. Related studies address knowledge-guided expert adaptation~\cite{fang2026moka}, few-shot text correction~\cite{feng2024protect}, multi-stage reasoning alignment~\cite{zhang2026lingualift}, data selection~\cite{dai2026deft}, local privacy protection~\cite{li2025rapt}, and catastrophic forgetting in continual fine-tuning~\cite{luo2025forgetting,chen2024bayesian}. Although these approaches reduce parameter and data budgets, computing exact gradients still caches layerwise activations across the frozen backbone.

Complementary compression methods lower static model memory via joint pruning and quantization~\cite{liu2025mixpq} or spectral-decoupled low-bit optimization~\cite{wei2026dsdllm}. These techniques compress the backbone weights and preserve orthogonality to activation-free gradient estimation during fine-tuning. Prompt-learning studies further show that downstream language adaptation concentrates within low-dimensional intrinsic task subspaces~\cite{qin2024intrinsicsubspace,zheng2024blackbox}. In forward-only fine-tuning, MeZO~\cite{malladi2023fine} and subsequent ZO studies~\cite{chen2024deepzero,Zhang2024ZOBenchMark} estimate the gradient from loss evaluations at two points. Each evaluation is a forward pass, so no activations are stored for backpropagation. Euclidean ZO variants reduce estimator variance via sparse coordinate selection~\cite{liu2024sparsemezo,curvzo2025}, diagonal curvature scaling~\cite{zhao2025hizoo}, batched or projected perturbations~\cite{dang2025fzoo,Mi2025PGAP_ProjGrad}, and candidate-pool screening~\cite{jin2026steering}, while low-rank extensions perturb either full-weight subspaces~\cite{chen2025lozo,Yu2025ZORandomSubspace} or coupled adapter factors~\cite{jin2024dfolora,yang2024Adazeta}.

During first-order training, Riemannian optimization preserves exact low-rank structure in tensor-train speech models~\cite{qi2023riemanntt} and matrix-manifold network adaptation~\cite{bian2025finding,zhangpilanci2024rplora,riemannion2025,mo2025loro}. Existing Riemannian ZO theories establish convergence on compact manifolds with retracted queries~\cite{li2022stochasticZO,li2023zorasa}, under coordinate-wise finite differences~\cite{he2024riemacc}, or via conformal metrics on open domains~\cite{ma2026structure}. Extrinsic finite-difference schemes also avoid query-point retractions via full tangent-basis evaluations on embedded submanifolds~\cite{taminiau2026riemannian}. However, coordinate-wise and tangent-basis differences require a number of evaluations proportional to the manifold dimension. Fixed-rank manifolds are noncompact, and retracted queries incur extra per-step SVD operations. Meanwhile, RoZO~\cite{rozo2026} combines coupled-factor tangent perturbations with transported momentum, adaptive preconditioning, and trust-region heuristics. By contrast, our method optimizes each update block directly on the fixed-rank manifold using a two-pass normalized tangent ZO estimator with provable nonconvex stationarity guarantees.

\vspace{-.2cm}
\subsection{Contributions}
\label{subsec:contributions}

Our main contributions are summarized as follows.

1) We develop ZO-RGD and its scalar-smoothed variant (ZO-RGD-S) for forward-only PEFT of LLMs. ZO-RGD combines the normalized tangent ZO estimator with a single truncated-SVD retraction per block to maintain exact fixed-rank updates at inference-level memory cost. ZO-RGD-S further smooths the shared finite-difference scalar across blocks through a single-scalar recursion without cross-tangent vector transport.

2) We establish theoretical guarantees for the normalized tangent ZO estimator and ZO-RGD on fixed-rank product manifolds. We prove that the estimator matches retracted queries to first order, remains invariant to factor parameterization, and aligns in expectation with the Riemannian gradient via tangent-space spherical isotropy. Combining these properties with a global second-order error bound for truncated-SVD retraction, we derive a nonconvex stationarity bound whose leading stochastic term scales with the manifold dimension.

3) We evaluate ZO-RGD and ZO-RGD-S across language models from 1.3 billion to 30 billion parameters on NLU, reading comprehension, and reasoning benchmarks. Both methods achieve competitive accuracy and convergence against forward-only baselines at inference-level memory cost. Controlled ablations confirm the role of tangent projection and perturbation normalization, while retracted manifold updates preserve exact rank-$r$ adapters throughout training.

\vspace{-.2cm}
\subsection{Organization and Notation}
\label{subsec:organization}

The remainder of this paper is organized as follows. Section~\ref{sec:method} reviews ZO gradient estimation based on function values at two points and fixed-rank manifold geometry, formulates the adaptation problem on a product manifold, and develops the normalized tangent ZO estimator, ZO-RGD, and ZO-RGD-S. Section~\ref{sec:theory} establishes the theoretical guarantees. Section~\ref{sec:experiments} reports the numerical experiments, and Section~\ref{sec:conclusion} concludes the paper.

Boldface lowercase letters ($\boldsymbol b,\boldsymbol z$) denote vectors and boldface uppercase letters ($\boldsymbol W,\boldsymbol Z$) denote matrices. The transpose, trace, and Frobenius inner product are denoted by $(\cdot)^\top$, $\operatorname{tr}(\cdot)$, and $\langle\boldsymbol G_1,\boldsymbol G_2\rangle_{\mathrm F}=\operatorname{tr}(\boldsymbol G_1^\top\boldsymbol G_2)$, respectively, with induced Frobenius norm $\|\boldsymbol G\|_{\mathrm F}=\sqrt{\langle\boldsymbol G,\boldsymbol G\rangle_{\mathrm F}}$. For vectors, $\langle\cdot,\cdot\rangle$ and $\|\cdot\|$ denote the Euclidean inner product and $\ell_2$ norm. On matrix--vector product spaces, the same unsubscripted symbols denote the standard inner product and induced norm. We write $\boldsymbol I_r$ for the $r\times r$ identity matrix and $\boldsymbol 0$ for a zero vector or matrix of conformable dimensions. Finally, $\mathbb E[\cdot]$ denotes expectation, $\sigma_i(\boldsymbol X)$ denotes the $i$-th largest singular value of $\boldsymbol X$, and $\|\boldsymbol X\|_2=\sigma_1(\boldsymbol X)$ denotes the spectral norm of $\boldsymbol X$.

\section{Proposed Method}
\label{sec:method}

\subsection{Zeroth-Order Optimization}
\label{subsec:zo_prelim}

Consider minimizing $f(\boldsymbol X)=\mathbb E_\xi[F(\boldsymbol X;\xi)]$ on $\mathbb R^{m\times n}$, where each minibatch loss $F(\cdot;\xi)$ is smooth. ZO methods estimate gradients from function evaluations alone. Given $\boldsymbol X$ and a minibatch $\xi$, let $\boldsymbol Z\in\mathbb R^{m\times n}$ have independent and identically distributed (i.i.d.) standard Gaussian $\mathcal N(0,1)$ entries. Evaluating the loss at two symmetric perturbations gives the ZO gradient estimator~\cite{Duchi2015RGDTIT,Nesterov2017RGE,shamir2017optimal,berahas2022comparison,balasubramanian2022zeroth}
\begin{equation}
    \widehat{\nabla}F(\boldsymbol X;\xi)
    =
    \frac{F(\boldsymbol X+\epsilon\boldsymbol Z;\xi)
    -F(\boldsymbol X-\epsilon\boldsymbol Z;\xi)}
    {2\epsilon}\,\boldsymbol Z,
    \label{eq:standard_rge}
\end{equation}
where $\epsilon>0$ is the finite-difference perturbation radius. With Euclidean gradient $\nabla F(\boldsymbol X;\xi)$, Taylor expansion around $\boldsymbol X$ shows that the ratio in~\eqref{eq:standard_rge} approximates $\langle\nabla F(\boldsymbol X;\xi),\boldsymbol Z\rangle_{\mathrm F}$ up to $\mathcal O(\epsilon^2)$. For any deterministic $\boldsymbol G\in\mathbb R^{m\times n}$, Gaussian isotropy gives $\mathbb E_{\boldsymbol Z}[\langle\boldsymbol G,\boldsymbol Z\rangle_{\mathrm F}\boldsymbol Z]=\boldsymbol G$, so $\mathbb E_{\boldsymbol Z}[\widehat{\nabla}F(\boldsymbol X;\xi)]=\nabla F(\boldsymbol X;\xi)+\mathcal O(\epsilon^2)$. With learning rate $\eta>0$, iteration $t$ updates $\boldsymbol X_{t+1}=\boldsymbol X_t-\eta\widehat{\nabla}F(\boldsymbol X_t;\xi_t)$. In LLM fine-tuning, MeZO~\cite{malladi2023fine} evaluates~\eqref{eq:standard_rge} with two forward passes per step. It regenerates $\boldsymbol Z$ in place from pseudorandom seeds without storing dense perturbation buffers.

\vspace{-.2cm}
\subsection{Fixed-Rank Manifold Geometry}
\label{subsec:manifold_prelim}

Forward-only ZO fine-tuning keeps training memory at the inference level, yet existing low-rank variants either accumulate high-rank updates in the full weight space or perturb coupled factors whose scaling is nonunique. To retain a compact low-rank update without this factor-scaling ambiguity, we optimize each additive update block directly on a fixed-rank matrix manifold with the pretrained weights frozen.

For dimensions $m\times n$ and rank $1\le r\le\min\{m,n\}$, the fixed-rank matrix manifold is defined as
\begin{equation}
    \mathcal M_r^{m\times n}
    \triangleq\{\boldsymbol X\in\mathbb R^{m\times n}\mid
    \operatorname{rank}(\boldsymbol X)=r\},
    \label{eq:fixed_rank_manifold}
\end{equation}
which we write as $\mathcal M_r$ when the dimensions are clear from the context. Every $\boldsymbol X\in\mathcal M_r$ admits a thin SVD $\boldsymbol X=\boldsymbol U\boldsymbol\Sigma\boldsymbol V^\top$, where $\boldsymbol U\in\mathbb R^{m\times r}$ and $\boldsymbol V\in\mathbb R^{n\times r}$ have orthonormal columns. The diagonal factor $\boldsymbol\Sigma\in\mathbb R^{r\times r}$ collects the positive singular values $\sigma_1(\boldsymbol X)\ge\cdots\ge\sigma_r(\boldsymbol X)>0$. The set $\mathcal M_r$ is a smooth embedded submanifold of $\mathbb R^{m\times n}$ with dimension $d_r\triangleq r(m+n-r)$~\cite{absil2008optimization,vandereycken2013low}. Its tangent space $T_{\boldsymbol X}\mathcal M_r$ at $\boldsymbol X$ consists of matrices of the form
\begin{equation}
    \boldsymbol Y
    =\boldsymbol U\boldsymbol M\boldsymbol V^\top
    +\boldsymbol U_p\boldsymbol V^\top
    +\boldsymbol U\boldsymbol V_p^\top,
    \label{eq:tangent_space}
\end{equation}
where $\boldsymbol M\in\mathbb R^{r\times r}$, while $\boldsymbol U_p\in\mathbb R^{m\times r}$ and $\boldsymbol V_p\in\mathbb R^{n\times r}$ satisfy $\boldsymbol U^\top\boldsymbol U_p=\boldsymbol 0$ and $\boldsymbol V^\top\boldsymbol V_p=\boldsymbol 0$.

Under the restricted Frobenius inner product, the orthogonal projection of an ambient matrix $\boldsymbol Z\in\mathbb R^{m\times n}$ onto $T_{\boldsymbol X}\mathcal M_r$ takes the closed form~\cite{absil2008optimization,vandereycken2013low,wei2016guarantees}
\begin{equation}
    P_{T_{\boldsymbol X}\mathcal M_r}(\boldsymbol Z)
    =\boldsymbol U\boldsymbol U^\top\boldsymbol Z
    +\boldsymbol Z\boldsymbol V\boldsymbol V^\top
    -\boldsymbol U\boldsymbol U^\top\boldsymbol Z\boldsymbol V\boldsymbol V^\top.
    \label{eq:orthogonal_projection}
\end{equation}
For a smooth function $f$ on $\mathbb R^{m\times n}$ with Euclidean gradient $\nabla f(\boldsymbol X)$, its Riemannian gradient on $\mathcal M_r$ is $\operatorname{grad}_{\mathcal M_r}f(\boldsymbol X)=P_{T_{\boldsymbol X}\mathcal M_r}(\nabla f(\boldsymbol X))$~\cite{absil2008optimization}. To map an ambient step $\boldsymbol X+\boldsymbol Y$ with $\boldsymbol Y\in T_{\boldsymbol X}\mathcal M_r$ back to $\mathcal M_r$, we use the rank-$r$ truncated-SVD retraction
\begin{equation}
    \mathcal R_{\boldsymbol X}(\boldsymbol Y)
    =\mathcal H_r(\boldsymbol X+\boldsymbol Y),
    \label{eq:retraction}
\end{equation}
where the truncated-SVD operator $\mathcal H_r$ retains at most $r$ leading singular triplets, minimizing the Frobenius distance over all matrices of rank at most $r$. Whenever $\operatorname{rank}(\boldsymbol X+\boldsymbol Y)\ge r$, $\mathcal H_r(\boldsymbol X+\boldsymbol Y)$ lies in $\mathcal M_r$ and gives the best rank-$r$ Frobenius approximation. Near $\boldsymbol 0\in T_{\boldsymbol X}\mathcal M_r$, $\mathcal R_{\boldsymbol X}$ is uniquely defined and forms a smooth second-order retraction onto $\mathcal M_r$~\cite{absil2008optimization,vandereycken2013low,bian2025finding}. In particular, every $\boldsymbol Y\in T_{\boldsymbol X}\mathcal M_r$ satisfies $\mathcal R_{\boldsymbol X}(\tau\boldsymbol Y)=\boldsymbol X+\tau\boldsymbol Y+\mathcal O(\tau^2)$ as $\tau\to0$.

Next, we formulate forward-only LLM adaptation on the resulting mixed product manifold, then derive a normalized tangent ZO gradient estimator that requires no query-point retractions, and finally develop ZO-RGD and ZO-RGD-S.

\vspace{-.2cm}
\subsection{Problem Formulation}
\label{subsec:problem_formulation}

Let $\mathcal S=\{1,\ldots,N\}$ index the $N$ adapted two-dimensional (2D) weight blocks of a pretrained LLM. During forward evaluation, block $\ell\in\mathcal S$ takes the additive form
\begin{equation}
    \boldsymbol W_\ell=\boldsymbol W_{0,\ell}+\Delta\boldsymbol W_\ell,
    \qquad
    \Delta\boldsymbol W_\ell\in\mathcal M_{r_\ell}^{m_\ell\times n_\ell},
    \label{eq:additive_weight}
\end{equation}
where $\boldsymbol W_{0,\ell}\in\mathbb R^{m_\ell\times n_\ell}$ is frozen and $\Delta\boldsymbol W_\ell$ is stored as a thin-SVD triple $(\boldsymbol U_\ell,\boldsymbol\Sigma_\ell,\boldsymbol V_\ell)$. Trainable one-dimensional (1D) biases and normalization scales are collected in $\boldsymbol b\in\mathbb R^{d_0}$, while all remaining parameters stay frozen. The trainable state then lies on the mixed product manifold
\begin{equation}
    \mathcal X=\mathcal M\times\mathbb R^{d_0},
    \qquad
    \mathcal M=\prod_{\ell\in\mathcal S}\mathcal M_{r_\ell}^{m_\ell\times n_\ell}
    \subset\mathcal E\triangleq\prod_{\ell\in\mathcal S}\mathbb R^{m_\ell\times n_\ell}.
    \label{eq:product_manifold}
\end{equation}
For a minibatch $\xi$, the task loss on the ambient space $\mathcal E\times\mathbb R^{d_0}$ is
\begin{equation}
    F(\Delta\boldsymbol W,\boldsymbol b;\xi)
    \triangleq\mathcal L(\boldsymbol W_0+\Delta\boldsymbol W,\boldsymbol b;\xi),
    \quad(\Delta\boldsymbol W,\boldsymbol b)\in\mathcal E\times\mathbb R^{d_0},
    \label{eq:ambient_loss}
\end{equation}
where $\boldsymbol W_0=(\boldsymbol W_{0,1},\ldots,\boldsymbol W_{0,N})$ and $\Delta\boldsymbol W=(\Delta\boldsymbol W_1,\ldots,\Delta\boldsymbol W_N)$ collect the frozen and update blocks, and $\mathcal L$ denotes the network loss, coupling all blocks through forward propagation. Forward-only fine-tuning is then formulated as
\begin{equation}
    \min_{(\Delta\boldsymbol W,\boldsymbol b)\in\mathcal X}
    f(\Delta\boldsymbol W,\boldsymbol b)
    \triangleq\mathbb E_\xi[F(\Delta\boldsymbol W,\boldsymbol b;\xi)].
    \label{eq:manifold_problem}
\end{equation}

\vspace{-.2cm}
\subsection{ZO Riemannian Gradient Estimation}
\label{ssec:tangent_estimation}

We first derive the estimator on a single block $\Delta\boldsymbol W_t\in\mathcal M_r$ of size $m\times n$ (omitting $\ell$ and holding the remaining blocks and $\boldsymbol b_t$ fixed in $F(\cdot;\xi_t)$), and then extend it to all blocks on $\mathcal X$.

\emph{Finite-difference scale mismatch.}
Sampling $\boldsymbol Z_t\in\mathbb R^{m\times n}$ with i.i.d.\ $\mathcal N(0,1)$ entries and projecting onto $T_{\Delta\boldsymbol W_t}\mathcal M_r$ via~\eqref{eq:orthogonal_projection} gives
\begin{equation}
    \widehat{\boldsymbol Z}_t
    =P_{T_{\Delta\boldsymbol W_t}\mathcal M_r}(\boldsymbol Z_t).
    \label{eq:tangent_projection_probe}
\end{equation}
Since $P_{T_{\Delta\boldsymbol W_t}\mathcal M_r}$ is the orthogonal projection operator onto a $d_r$-dimensional subspace, $\|\boldsymbol Z_t\|_{\mathrm F}^2$ and $\|\widehat{\boldsymbol Z}_t\|_{\mathrm F}^2$ (given $\Delta\boldsymbol W_t$) follow chi-squared distributions $\chi^2_{mn}$ and $\chi^2_{d_r}$, respectively, giving
\begin{equation}
    \mathbb E\bigl[\|\boldsymbol Z_t\|_{\mathrm F}^2\bigr]=mn,
    \qquad
    \mathbb E\bigl[\|\widehat{\boldsymbol Z}_t\|_{\mathrm F}^2\mid\Delta\boldsymbol W_t\bigr]=d_r.
    \label{eq:norm_shrinkage}
\end{equation}
When $r\ll\min\{m,n\}$, $d_r\ll mn$ shrinks the perturbation norm by $\sqrt{d_r/(mn)}$, coupling the effective query radius $\epsilon\|\widehat{\boldsymbol Z}_t\|_{\mathrm F}$ to $(m,n,r)$.

\emph{Normalized tangent ZO estimator.}
To match the root-mean-square ambient scale across ranks and layer shapes, we normalize $\widehat{\boldsymbol Z}_t$ as
\begin{equation}
    \widetilde{\boldsymbol Z}_t
    =\sqrt{mn}\,
    \frac{\widehat{\boldsymbol Z}_t}{\|\widehat{\boldsymbol Z}_t\|_{\mathrm F}},
    \label{eq:zo_norm}
\end{equation}
where $\|\widehat{\boldsymbol Z}_t\|_{\mathrm F}>0$ almost surely since $d_r\ge 1$, so that $\|\widetilde{\boldsymbol Z}_t\|_{\mathrm F}=\sqrt{mn}$. Two forward evaluations on minibatch $\xi_t$ at $\Delta\boldsymbol W_t\pm\epsilon\widetilde{\boldsymbol Z}_t$ give the finite-difference directional derivative and tangent estimator
\begin{align}
    \hat s_t
    &=\frac{
    F(\Delta\boldsymbol W_t+\epsilon\widetilde{\boldsymbol Z}_t;\xi_t)
    -F(\Delta\boldsymbol W_t-\epsilon\widetilde{\boldsymbol Z}_t;\xi_t)}
    {2\epsilon},
    \label{eq:zo_scalar}\\
    \widehat{\boldsymbol G}_t
    &=\hat s_t\widetilde{\boldsymbol Z}_t
    \in T_{\Delta\boldsymbol W_t}\mathcal M_r.
    \label{eq:zo_grad}
\end{align}
On the product space $\mathcal X$ in~\eqref{eq:manifold_problem}, drawing independent block directions $\widetilde{\boldsymbol Z}_t=(\widetilde{\boldsymbol Z}_{t,1},\ldots,\widetilde{\boldsymbol Z}_{t,N})$ via~\eqref{eq:zo_norm} together with $\boldsymbol z_{0,t}\sim\mathcal N(\boldsymbol 0,\boldsymbol I_{d_0})$ yields the shared scalar
\begin{align}
    \hat s_t
    =\frac{1}{2\epsilon}\Bigl[
    &F(\Delta\boldsymbol W_t+\epsilon\widetilde{\boldsymbol Z}_t,\boldsymbol b_t+\epsilon\boldsymbol z_{0,t};\xi_t)
    \nonumber\\
    &-F(\Delta\boldsymbol W_t-\epsilon\widetilde{\boldsymbol Z}_t,\boldsymbol b_t-\epsilon\boldsymbol z_{0,t};\xi_t)\Bigr],
    \label{eq:joint_zo_scalar}
\end{align}
and block estimators $\widehat{\boldsymbol G}_{t,\ell}=\hat s_t\widetilde{\boldsymbol Z}_{t,\ell}\in T_{\Delta\boldsymbol W_{t,\ell}}\mathcal M_{r_\ell}$ for $\ell\in\mathcal S$, with $\hat{\boldsymbol g}_{0,t}=\hat s_t\boldsymbol z_{0,t}$ for $\boldsymbol b_t$. Evaluating~\eqref{eq:ambient_loss} along the ambient tangent line $\Delta\boldsymbol W_{t,\ell}\pm\epsilon\widetilde{\boldsymbol Z}_{t,\ell}$ (rank at most $2r_\ell$) avoids two SVD retractions per block. Propositions~\ref{prop:probe_equiv} and~\ref{prop:tangent_sphere} show that this rule preserves first-order accuracy and Riemannian gradient alignment.

\vspace{-.2cm}
\subsection{ZO Riemannian Optimization Algorithms}
\label{sec:algorithms}

As illustrated in Fig.~\ref{fig:method_overview}, combining the tangent estimator with the truncated-SVD retraction~\eqref{eq:retraction} yields ZO-RGD and its smoothed variant ZO-RGD-S.

\emph{ZO-RGD.}
Given learning rate $\eta>0$ and $\hat s_t$ in~\eqref{eq:joint_zo_scalar}, ZO-RGD updates $\boldsymbol x_t\triangleq(\Delta\boldsymbol W_t,\boldsymbol b_t)\in\mathcal X$ across all blocks $\ell\in\mathcal S$ in parallel via
\begin{align}
    \Delta\boldsymbol W_{t+1,\ell}
    &=\mathcal R_{\Delta\boldsymbol W_{t,\ell}}
    \bigl(-\eta\widehat{\boldsymbol G}_{t,\ell}\bigr),
    \qquad \ell\in\mathcal S,
    \label{eq:zo_rgd_update}\\
    \boldsymbol b_{t+1}
    &=\boldsymbol b_t-\eta\hat{\boldsymbol g}_{0,t}.
    \label{eq:zo_rgd_1d}
\end{align}
In~\eqref{eq:orthogonal_projection}, $P_{T_{\Delta\boldsymbol W_{t,\ell}}\mathcal M_{r_\ell}}$ depends only on $\boldsymbol U_\ell\boldsymbol U_\ell^\top$ and $\boldsymbol V_\ell\boldsymbol V_\ell^\top$, while $\mathcal R_{\Delta\boldsymbol W_{t,\ell}}$ in~\eqref{eq:retraction} depends only on the displaced matrix. Hence,~\eqref{eq:zo_rgd_update} is invariant to the choice of thin-SVD factors. Since $\boldsymbol 0\notin\mathcal M_{r_\ell}^{m_\ell\times n_\ell}$ for $r_\ell\ge1$, we initialize $\Delta\boldsymbol W_{0,\ell}=\sigma_{\mathrm{init}}\boldsymbol U_{0,\ell}\boldsymbol V_{0,\ell}^\top$ with $\sigma_{\mathrm{init}}=10^{-6}$. Here, $\boldsymbol U_{0,\ell}\in\mathbb R^{m_\ell\times r_\ell}$ and $\boldsymbol V_{0,\ell}\in\mathbb R^{n_\ell\times r_\ell}$ have orthonormal columns obtained from standard Gaussian matrices via thin orthogonal--triangular (QR) factorization. Algorithm~\ref{alg:zo_rgd} summarizes ZO-RGD.

\begin{algorithm}[!t]
\caption{Zeroth-Order Riemannian Gradient Descent (ZO-RGD)}
\label{alg:zo_rgd}
\begin{algorithmic}[1]
\State \textbf{Input:} Frozen weights $\boldsymbol W_0$, learning rate $\eta$, perturbation radius $\epsilon$, iterations $T$.
\State \textbf{Initialize:} $\Delta\boldsymbol W_0\in\mathcal M$, $\boldsymbol b_0\in\mathbb R^{d_0}$.
\For{$t=0,1,\ldots,T-1$}
    \State Sample minibatch $\xi_t$, $\{\boldsymbol Z_{t,\ell}\}_{\ell\in\mathcal S}$, and $\boldsymbol z_{0,t}\sim\mathcal N(\boldsymbol 0,\boldsymbol I_{d_0})$
    \ForAll{$\ell\in\mathcal S$ in parallel}
        \State $\widehat{\boldsymbol Z}_{t,\ell} \gets P_{T_{\Delta\boldsymbol W_{t,\ell}}\mathcal M_{r_\ell}}(\boldsymbol Z_{t,\ell})$ \Comment{tangent projection}
        \State $\widetilde{\boldsymbol Z}_{t,\ell} \gets \sqrt{m_\ell n_\ell}\,\widehat{\boldsymbol Z}_{t,\ell}/\|\widehat{\boldsymbol Z}_{t,\ell}\|_{\mathrm F}$ \Comment{normalization}
    \EndFor
    \State $\hat s_t \gets \bigl[F(\Delta\boldsymbol W_t+\epsilon\widetilde{\boldsymbol Z}_t,\boldsymbol b_t+\epsilon\boldsymbol z_{0,t};\xi_t)-F(\Delta\boldsymbol W_t-\epsilon\widetilde{\boldsymbol Z}_t,\boldsymbol b_t-\epsilon\boldsymbol z_{0,t};\xi_t)\bigr]/(2\epsilon)$
    \ForAll{$\ell\in\mathcal S$ in parallel}
        \State $\widehat{\boldsymbol G}_{t,\ell} \gets \hat s_t\widetilde{\boldsymbol Z}_{t,\ell}$ \Comment{Riemannian ZO gradient}
        \State $\Delta\boldsymbol W_{t+1,\ell} \gets \mathcal R_{\Delta\boldsymbol W_{t,\ell}}(-\eta\widehat{\boldsymbol G}_{t,\ell})$ \Comment{SVD retraction}
    \EndFor
    \State $\hat{\boldsymbol g}_{0,t} \gets \hat s_t\boldsymbol z_{0,t}$;\; $\boldsymbol b_{t+1} \gets \boldsymbol b_t-\eta\hat{\boldsymbol g}_{0,t}$ \Comment{Euclidean 1D step}
\EndFor
\State \textbf{Output:} $\Delta\boldsymbol W_T\in\mathcal M$, $\boldsymbol b_T\in\mathbb R^{d_0}$.
\end{algorithmic}
\end{algorithm}

\emph{ZO-RGD-S.}
Under the shared scalar $\hat s_t$, block $\ell$ takes a step of norm $\|\eta\widehat{\boldsymbol G}_{t,\ell}\|_{\mathrm F}=\eta|\hat s_t|\sqrt{m_\ell n_\ell}$, which is largest on the leading token-embedding matrix. ZO-RGD-S retains the same two forward passes and smooths $\hat s_t$ across a fixed block ordering $\ell_1,\ldots,\ell_N$ of $\mathcal S$ with $\beta\in[0,1)$ via
\begin{equation}
    \bar s_{t,k}
    =\beta \bar s_{t,k-1}+(1-\beta)\hat s_t,
    \qquad k=1,\ldots,N,
    \label{eq:scalar_recursive_smoothing}
\end{equation}
initialized at $\bar s_{0,0}=\hat s_0$ and carried across steps as $\bar s_{t,0}=\bar s_{t-1,N}$ for $t\ge1$, while $\boldsymbol b_t$ updates via~\eqref{eq:zo_rgd_1d}. Block $\ell_k$ then updates via
\begin{align}
    \widehat{\boldsymbol G}^{\mathrm S}_{t,\ell_k}
    &=\bar s_{t,k}\widetilde{\boldsymbol Z}_{t,\ell_k},
    \label{eq:zo_grad_smoothed}\\
    \Delta\boldsymbol W_{t+1,\ell_k}
    &=\mathcal R_{\Delta\boldsymbol W_{t,\ell_k}}
    \bigl(-\eta\widehat{\boldsymbol G}^{\mathrm S}_{t,\ell_k}\bigr),
    \label{eq:zo_rgds_update}
\end{align}
which reduces to ZO-RGD at $t=0$ or $\beta=0$. Unrolling~\eqref{eq:scalar_recursive_smoothing} for $t\ge1$ gives
\begin{equation}
    \bar s_{t,k}=(1-\beta^k)\hat s_t+\beta^k\bar s_{t-1,N}.
    \label{eq:scalar_closed_form}
\end{equation}
Let $\mathcal F_t$ denote the $\sigma$-algebra generated by $\boldsymbol x_0$ and all minibatches and perturbations prior to iteration $t$. Conditioned on $\mathcal F_t$, $\bar s_{t-1,N}$ is deterministic and $\mathbb E[\widetilde{\boldsymbol Z}_{t,\ell_k}\mid\mathcal F_t]=\boldsymbol 0$, so for $t\ge1$,~\eqref{eq:zo_grad_smoothed} and~\eqref{eq:scalar_closed_form} yield
\begin{equation}
    \mathbb E\bigl[\widehat{\boldsymbol G}^{\mathrm S}_{t,\ell_k}\mid\mathcal F_t\bigr]
    =(1-\beta^k)\,\mathbb E\bigl[\widehat{\boldsymbol G}_{t,\ell_k}\mid\mathcal F_t\bigr].
    \label{eq:zo_rgds_expectation}
\end{equation}
Thus, $\beta\in(0,1)$ ($\beta=0.9$ in Section~\ref{sec:experiments}) scales the expected step size geometrically from $(1-\beta)\eta$ toward $\eta$ along $\ell_1,\ldots,\ell_N$ without vector transport. Meanwhile, the carried term $\beta^k\bar s_{t-1,N}\widetilde{\boldsymbol Z}_{t,\ell_k}$ has zero conditional mean and adds isotropic tangent variance $\beta^{2k}\bar s_{t-1,N}^2 m_{\ell_k}n_{\ell_k}$. Algorithm~\ref{alg:zo_rgds} summarizes ZO-RGD-S.

\begin{algorithm}[!t]
\caption{ZO-RGD with smoothed scalar (ZO-RGD-S)}
\label{alg:zo_rgds}
\begin{algorithmic}[1]
\State \textbf{Input:} Frozen weights $\boldsymbol W_0$, block ordering $\ell_1,\ldots,\ell_N$, learning rate $\eta$, radius $\epsilon$, smoothing $\beta\in[0,1)$, iterations $T$.
\State \textbf{Initialize:} $\Delta\boldsymbol W_0\in\mathcal M$, $\boldsymbol b_0\in\mathbb R^{d_0}$.
\For{$t=0,1,\ldots,T-1$}
    \State Sample minibatch $\xi_t$, construct $\{\widetilde{\boldsymbol Z}_{t,\ell_k}\}_{k=1}^N$ and $\boldsymbol z_{0,t}$, and evaluate $\hat s_t$ as in Algorithm~\ref{alg:zo_rgd}
    \If{$t=0$}
        \State $\bar s_{0,0} \gets \hat s_0$ \Comment{initialize scalar state}
    \Else
        \State $\bar s_{t,0} \gets \bar s_{t-1,N}$ \Comment{carry across iterations}
    \EndIf
    \For{$k=1,\ldots,N$}
        \State $\bar s_{t,k} \gets \beta \bar s_{t,k-1} + (1-\beta)\hat s_t$ \Comment{scalar recursion}
        \State $\widehat{\boldsymbol G}^{\mathrm S}_{t,\ell_k} \gets \bar s_{t,k}\widetilde{\boldsymbol Z}_{t,\ell_k}$ \Comment{smoothed estimator}
        \State $\Delta\boldsymbol W_{t+1,\ell_k} \gets \mathcal R_{\Delta\boldsymbol W_{t,\ell_k}}(-\eta\widehat{\boldsymbol G}^{\mathrm S}_{t,\ell_k})$ \Comment{retraction}
    \EndFor
    \State $\hat{\boldsymbol g}_{0,t} \gets \hat s_t\boldsymbol z_{0,t}$;\; $\boldsymbol b_{t+1} \gets \boldsymbol b_t-\eta\hat{\boldsymbol g}_{0,t}$
\EndFor
\State \textbf{Output:} $\Delta\boldsymbol W_T\in\mathcal M$, $\boldsymbol b_T\in\mathbb R^{d_0}$.
\end{algorithmic}
\end{algorithm}

\emph{Blockwise implementation and memory cost.}
Storing $(\boldsymbol U_\ell,\boldsymbol\Sigma_\ell,\boldsymbol V_\ell)$ requires $r_\ell(m_\ell+n_\ell+1)$ parameters per block, plus one scalar across all blocks for ZO-RGD-S. Each Gaussian matrix $\boldsymbol Z_{t,\ell}$ is regenerated on the fly from per-step seeds~\cite{malladi2023fine,chen2025lozo}. Per block, sampling and projecting $\boldsymbol Z_{t,\ell}$ costs $\mathcal O(m_\ell n_\ell r_\ell)$ operations. Using the tangent form~\eqref{eq:tangent_space}, evaluating $\mathcal H_{r_\ell}(\Delta\boldsymbol W_{t,\ell}-\eta\widehat{\boldsymbol G}_{t,\ell})$ via thin QR factorizations and a $2r_\ell\times 2r_\ell$ SVD costs $\mathcal O((m_\ell+n_\ell)r_\ell^2+r_\ell^3)$ operations.

\section{Theoretical Results}
\label{sec:theory}

\subsection{Directional Properties of the Tangent Estimator}
\label{ssec:estimator_properties}

\begin{proposition}[Ambient and retracted queries]
\label{prop:probe_equiv}
Let $F(\cdot;\xi_t)$ be differentiable near $\Delta\boldsymbol W_t\in\mathcal M_r$, and let $\mathcal R_{\Delta\boldsymbol W_t}(\tau\boldsymbol Y)=\Delta\boldsymbol W_t+\tau\boldsymbol Y+\mathcal O(\tau^2)$ as $\tau\to0$ for all $\boldsymbol Y\in T_{\Delta\boldsymbol W_t}\mathcal M_r$. For sufficiently small $\epsilon>0$, define the retracted finite difference
\begin{equation}
    \hat s_t^{\mathrm{retr}}
    \triangleq\frac{
    F(\mathcal R_{\Delta\boldsymbol W_t}(\epsilon\widetilde{\boldsymbol Z}_t);\xi_t)
    -F(\mathcal R_{\Delta\boldsymbol W_t}(-\epsilon\widetilde{\boldsymbol Z}_t);\xi_t)}
    {2\epsilon}.
    \label{eq:retr_scalar}
\end{equation}
Then $\hat s_t$ in~\eqref{eq:zo_scalar} satisfies $\hat s_t=\hat s_t^{\mathrm{retr}}+o(1)$ as $\epsilon\to0$.
\end{proposition}

\begin{proof}
First-order differentiability of $F(\cdot;\xi_t)$ at $\Delta\boldsymbol W_t$ implies that the ambient tangent-line difference~\eqref{eq:zo_scalar} satisfies
\begin{equation}
    \hat s_t
    =\bigl\langle\nabla F(\Delta\boldsymbol W_t;\xi_t),\widetilde{\boldsymbol Z}_t\bigr\rangle_{\mathrm F}+o(1).
\end{equation}
By the first-order tangency property of the retraction $\mathcal R_{\Delta\boldsymbol W_t}$, we have $\mathcal R_{\Delta\boldsymbol W_t}(\pm\epsilon\widetilde{\boldsymbol Z}_t)=\Delta\boldsymbol W_t\pm\epsilon\widetilde{\boldsymbol Z}_t+\mathcal O(\epsilon^2)$. Substituting this expansion into~\eqref{eq:retr_scalar} and using the differentiability of $F(\cdot;\xi_t)$ again yields
\begin{equation}
    \hat s_t^{\mathrm{retr}}
    =\bigl\langle\nabla F(\Delta\boldsymbol W_t;\xi_t),\widetilde{\boldsymbol Z}_t\bigr\rangle_{\mathrm F}+o(1),
\end{equation}
which establishes $\hat s_t=\hat s_t^{\mathrm{retr}}+o(1)$ as $\epsilon\to0$.
\end{proof}

Proposition~\ref{prop:probe_equiv} establishes first-order equivalence between the ambient query~\eqref{eq:zo_scalar} and the retracted query~\eqref{eq:retr_scalar} without two extra SVD retractions per block. Deterministic extrinsic finite-difference methods also evaluate ambient tangent lines on embedded submanifolds~\cite{taminiau2026riemannian}, whereas~\eqref{eq:zo_scalar}--\eqref{eq:zo_grad} use two stochastic passes without constructing a $d_r$-element tangent basis. At finite $\epsilon>0$, the ambient and retracted rules query distinct points, and all subsequent bounds analyze $\hat s_t$ in~\eqref{eq:zo_scalar} directly.

\begin{proposition}[Expected gradient alignment]
\label{prop:tangent_sphere}
Assume $F(\cdot;\xi_t)$ has a locally Lipschitz Hessian near $\Delta\boldsymbol W_t\in\mathcal M_r$. Under the Frobenius metric on $\mathcal M_r$, $\widehat{\boldsymbol G}_t$ in~\eqref{eq:zo_grad} satisfies
\begin{equation}
    \mathbb E_{\boldsymbol Z_t}
    [\widehat{\boldsymbol G}_t\mid\Delta\boldsymbol W_t,\xi_t]
    =\frac{mn}{d_r}\,
    \operatorname{grad}_{\mathcal M_r}F(\Delta\boldsymbol W_t;\xi_t)
    +\mathcal O(\epsilon^2)
    \label{eq:prop_expected_alignment}
\end{equation}
as $\epsilon\to0$. When differentiation commutes with $\mathbb E_{\xi_t}[\cdot]$, taking expectation over $\xi_t$ replaces $F(\cdot;\xi_t)$ with $f(\cdot)$ in~\eqref{eq:prop_expected_alignment}.
\end{proposition}

\begin{proof}
We write the estimator as a rank-one quadratic form in a direction uniform on the unit sphere of $T_t\triangleq T_{\Delta\boldsymbol W_t}\mathcal M_r$ (dimension $d_r=r(m+n-r)$) and apply spherical isotropy. Since $\widehat{\boldsymbol Z}_t=P_{T_t}(\boldsymbol Z_t)$ orthogonally projects an isotropic standard Gaussian matrix onto $T_t$, $\boldsymbol u_t\triangleq\widehat{\boldsymbol Z}_t/\|\widehat{\boldsymbol Z}_t\|_{\mathrm F}$ is uniform on the unit sphere of $T_t$, with $\widetilde{\boldsymbol Z}_t=\sqrt{mn}\,\boldsymbol u_t$. Expanding $F(\cdot;\xi_t)$ around $\Delta\boldsymbol W_t$ with a locally Lipschitz Hessian cancels the second-order terms in~\eqref{eq:zo_scalar}. Since $\|\widetilde{\boldsymbol Z}_t\|_{\mathrm F}=\sqrt{mn}$, the remainder is uniform in $\boldsymbol Z_t$ and gives
\begin{equation}
    \hat s_t
    =\bigl\langle\nabla F(\Delta\boldsymbol W_t;\xi_t),\widetilde{\boldsymbol Z}_t\bigr\rangle_{\mathrm F}+\mathcal O(\epsilon^2).
\end{equation}
Write $\boldsymbol g_t\triangleq\operatorname{grad}_{\mathcal M_r}F(\Delta\boldsymbol W_t;\xi_t)=P_{T_t}\bigl(\nabla F(\Delta\boldsymbol W_t;\xi_t)\bigr)$. Since $\boldsymbol u_t\in T_t$ and $P_{T_t}$ is self-adjoint, we have $\langle\nabla F(\Delta\boldsymbol W_t;\xi_t),\boldsymbol u_t\rangle_{\mathrm F}=\langle\boldsymbol g_t,\boldsymbol u_t\rangle_{\mathrm F}$, so that
\begin{equation}
    \widehat{\boldsymbol G}_t
    =\hat s_t\widetilde{\boldsymbol Z}_t
    =mn\,\langle\boldsymbol g_t,\boldsymbol u_t\rangle_{\mathrm F}\boldsymbol u_t+\mathcal O(\epsilon^2).
\end{equation}
By spherical isotropy on the $d_r$-dimensional subspace $T_t$, the second-moment operator satisfies $\mathbb E_{\boldsymbol Z_t}[\langle\boldsymbol g_t,\boldsymbol u_t\rangle_{\mathrm F}\boldsymbol u_t\mid\Delta\boldsymbol W_t,\xi_t]=(1/d_r)\boldsymbol g_t$, which yields~\eqref{eq:prop_expected_alignment}. Taking expectation over $\xi_t$ gives the corresponding identity for $f$.
\end{proof}

Here, the dimension ratio $mn/d_r\ge1$ depends only on $(m,n,r)$, and the projector $P_{T_{\Delta\boldsymbol W_t}\mathcal M_r}$ depends on $\Delta\boldsymbol W_t$ solely through its column and row spaces. By contrast, isotropic factor-induced tangent perturbations of the form in RoZO~\cite{rozo2026} induce a factor-dependent metric.

\begin{proposition}[Mean step of factor-induced tangent directions]
\label{prop:factor_probe}
Let $\boldsymbol X=\boldsymbol B\boldsymbol A\in\mathcal M_r$ with $\boldsymbol B\in\mathbb R^{m\times r}$ and $\boldsymbol A\in\mathbb R^{r\times n}$. Draw independent $\Delta\boldsymbol B\in\mathbb R^{m\times r}$ and $\Delta\boldsymbol A\in\mathbb R^{r\times n}$ with i.i.d.\ $\mathcal N(0,1)$ entries, and set $\boldsymbol\Delta=\boldsymbol B\Delta\boldsymbol A+\Delta\boldsymbol B\boldsymbol A$. Then $\boldsymbol\Delta\in T_{\boldsymbol X}\mathcal M_r$, and every $\boldsymbol G\in\mathbb R^{m\times n}$ satisfies
\begin{equation}
    \mathbb E\bigl[\langle\boldsymbol G,\boldsymbol\Delta\rangle_{\mathrm F}\boldsymbol\Delta\bigr]
    =\boldsymbol B\boldsymbol B^\top\boldsymbol G+\boldsymbol G\boldsymbol A^\top\boldsymbol A.
    \label{eq:factor_mean_step}
\end{equation}
\end{proposition}

\begin{proof}
For a thin SVD $\boldsymbol X=\boldsymbol U\boldsymbol\Sigma\boldsymbol V^\top$, full rank of $\boldsymbol B$ and $\boldsymbol A$ gives $\boldsymbol U\boldsymbol U^\top\boldsymbol B=\boldsymbol B$ and $\boldsymbol A\boldsymbol V\boldsymbol V^\top=\boldsymbol A$. Substituting $\boldsymbol\Delta$ into~\eqref{eq:orthogonal_projection} yields $P_{T_{\boldsymbol X}\mathcal M_r}(\boldsymbol\Delta)=\boldsymbol\Delta$, so $\boldsymbol\Delta\in T_{\boldsymbol X}\mathcal M_r$. Writing $\langle\boldsymbol G,\boldsymbol\Delta\rangle_{\mathrm F}=\langle\boldsymbol B^\top\boldsymbol G,\Delta\boldsymbol A\rangle_{\mathrm F}+\langle\boldsymbol G\boldsymbol A^\top,\Delta\boldsymbol B\rangle_{\mathrm F}$ and using the independence and standard Gaussian isotropy of $\Delta\boldsymbol A$ and $\Delta\boldsymbol B$ gives~\eqref{eq:factor_mean_step}.
\end{proof}

Under any invertible reparameterization $(\boldsymbol B\boldsymbol Q,\boldsymbol Q^{-1}\boldsymbol A)$ of $\boldsymbol X$,~\eqref{eq:factor_mean_step} changes whenever $\boldsymbol Q\boldsymbol Q^\top\neq\boldsymbol I_r$. Even for balanced factors $\boldsymbol B=\boldsymbol U\boldsymbol\Sigma^{1/2}$ and $\boldsymbol A=\boldsymbol\Sigma^{1/2}\boldsymbol V^\top$,~\eqref{eq:factor_mean_step} becomes $\boldsymbol U\boldsymbol\Sigma\boldsymbol U^\top\boldsymbol G+\boldsymbol G\boldsymbol V\boldsymbol\Sigma\boldsymbol V^\top$ and weights tangent directions by $\boldsymbol\Sigma$. In contrast, the directional covariance operator in~\eqref{eq:prop_expected_alignment} is invariant to both factor scaling and singular-value magnitudes.

\vspace{-.2cm}
\subsection{Stationarity Analysis of ZO-RGD}
\label{ssec:stationarity}

Write $\mathcal M_\ell=\mathcal M_{r_\ell}^{m_\ell\times n_\ell}$ with $m_\ell,n_\ell\ge 2r_\ell$ for $\ell\in\mathcal S$, and expand the ZO-RGD iterate as $\boldsymbol x_t=(\Delta\boldsymbol W_{t,1},\ldots,\Delta\boldsymbol W_{t,N},\boldsymbol b_t)\in\mathcal X$. Let $\boldsymbol\zeta_t=(\widetilde{\boldsymbol Z}_{t,1},\ldots,\widetilde{\boldsymbol Z}_{t,N},\boldsymbol z_{0,t})\in T_{\boldsymbol x_t}\mathcal X$ be the joint perturbation direction, so $\widehat{\boldsymbol G}_t=\hat s_t\boldsymbol\zeta_t$. Defining the product retraction $\mathcal R_{\boldsymbol x_t}$ by~\eqref{eq:zo_rgd_update}--\eqref{eq:zo_rgd_1d} gives $\boldsymbol x_{t+1}=\mathcal R_{\boldsymbol x_t}(-\eta\widehat{\boldsymbol G}_t)$. Denote by $\nabla_\ell f$ and $\nabla_0 f$ the Euclidean partial gradients of $f$ with respect to $\Delta\boldsymbol W_\ell$ and $\boldsymbol b$, respectively, with $\operatorname{grad}_{\mathcal M_\ell}f=P_{T_{\Delta\boldsymbol W_\ell}\mathcal M_\ell}(\nabla_\ell f)$. Set $\operatorname{grad}f\triangleq(\operatorname{grad}_{\mathcal M_1}f,\ldots,\operatorname{grad}_{\mathcal M_N}f,\nabla_0 f)$, $\bar D\triangleq\sum_{\ell=1}^N m_\ell n_\ell+d_0$, and $c_\ell\triangleq m_\ell n_\ell/d_{r_\ell}\ge1$ from~\eqref{eq:prop_expected_alignment}. Defining $\boldsymbol D(\boldsymbol v_1,\ldots,\boldsymbol v_N,\boldsymbol v_0)\triangleq(c_1\boldsymbol v_1,\ldots,c_N\boldsymbol v_N,\boldsymbol v_0)$ on $T_{\boldsymbol x_t}\mathcal X$, the weighted gradient norm is
\begin{equation}
    \|\operatorname{grad}f(\boldsymbol x_t)\|_{\boldsymbol D}^2
    \triangleq\sum_{\ell=1}^N c_\ell\|\operatorname{grad}_{\mathcal M_\ell}f(\boldsymbol x_t)\|_{\mathrm F}^2+\|\nabla_0 f(\boldsymbol x_t)\|^2.
    \label{eq:weighted_norm_main}
\end{equation}

Since $\{\boldsymbol x_t\}$ is random through $(\xi_t,\boldsymbol\zeta_t)$, fix $t_0\ge0$ and let $\mathcal X_{\mathrm{reg}}\subset\mathcal X$ be a subset containing $\{\boldsymbol x_t\}_{t\ge t_0}$ with probability one. Let $\mathcal X'_{\mathrm{reg}}$ collect $\mathcal X_{\mathrm{reg}}$ together with all realized query segments $\{\boldsymbol x_t+\tau\boldsymbol\zeta_t\mid |\tau|\le\epsilon\}$, step segments $\{\boldsymbol x_t-\tau\eta\widehat{\boldsymbol G}_t\mid\tau\in[0,1]\}$, and retraction segments between $\boldsymbol x_t-\eta\widehat{\boldsymbol G}_t$ and $\boldsymbol x_{t+1}$ for $t\ge t_0$. Because $\mathcal M_\ell$ is nonclosed, the second-order error of~\eqref{eq:retraction} grows if $\sigma_{r_\ell}(\Delta\boldsymbol W_{t,\ell})\to 0$. Assumption~\ref{ass:A} keeps the trajectory away from this boundary, while Assumptions~\ref{ass:B} and~\ref{ass:C} state standard smoothness and minibatch-variance conditions~\cite{ghadimi2013stochastic,Nesterov2017RGE}.

\begin{assumption}[Trajectory regularity]
\label{ass:A}
There exist constants $\sigma_{\min}>0$, $G_\infty<\infty$, and $f_*>-\infty$ such that, with probability one, (i) $\sigma_{r_\ell}(\Delta\boldsymbol W_{t,\ell})\ge\sigma_{\min}$ for all $\ell\in\mathcal S$ and $t\ge t_0$, (ii) $\|\nabla f(\boldsymbol x)\|\le G_\infty$ on $\mathcal X'_{\mathrm{reg}}$, and (iii) $f(\boldsymbol x)\ge f_*$ on $\mathcal X_{\mathrm{reg}}$.
\end{assumption}

\begin{assumption}[Smoothness]
\label{ass:B}
There exist constants $L>0$ and $\rho_3\ge0$ such that, for every $\xi$, $F(\cdot;\xi)$ is differentiable with an $L$-Lipschitz gradient on $\mathcal X'_{\mathrm{reg}}$. Along each realized query segment, $\phi(\tau)\triangleq F(\boldsymbol x+\tau\boldsymbol v;\xi)$ is thrice differentiable with $|\phi'''(\tau)|\le\rho_3\|\boldsymbol v\|^3$.
\end{assumption}

\begin{assumption}[Stochastic oracle]
\label{ass:C}
For all $\boldsymbol x\in\mathcal X'_{\mathrm{reg}}$, $\mathbb E_\xi[\nabla F(\boldsymbol x;\xi)]=\nabla f(\boldsymbol x)$, and both evaluations in~\eqref{eq:joint_zo_scalar} share $\xi_t$. Moreover, there exist constants $\sigma_\ell,\sigma_0\ge0$ such that, on $\mathcal X_{\mathrm{reg}}$, $\mathbb E_\xi\|\operatorname{grad}_{\mathcal M_\ell}F(\boldsymbol x;\xi)-\operatorname{grad}_{\mathcal M_\ell}f(\boldsymbol x)\|_{\mathrm F}^2\le\sigma_\ell^2$ for $\ell\in\mathcal S$ and $\mathbb E_\xi\|\nabla_0 F(\boldsymbol x;\xi)-\nabla_0 f(\boldsymbol x)\|^2\le\sigma_0^2$.
\end{assumption}

In our completed instrumented runs, $\min_\ell\sigma_{r_\ell}(\Delta\boldsymbol W_{t,\ell})>0.08$ from step $500$ onward (Section~\ref{ssec:ablation}), motivating $t_0$ in Assumption~\ref{ass:A}(i). Assumption~\ref{ass:B} bounds the third-order remainder of~\eqref{eq:joint_zo_scalar}, and Assumption~\ref{ass:C} enters through $\sigma_D^2$ below. Let $\Delta_0>0$ satisfy $\mathbb E[f(\boldsymbol x_{t_0})]-f_*\le\Delta_0$, and define
\begin{equation}
    L_R\triangleq L+\frac{4G_\infty}{\sigma_{\min}},
    \qquad
    \sigma_D^2\triangleq\sum_{\ell=1}^N c_\ell\sigma_\ell^2+\sigma_0^2.
    \label{eq:lr_sigmad}
\end{equation}

\begin{theorem}[Nonconvex stationarity bound for ZO-RGD]
\label{thm:convergence}
Suppose $m_\ell,n_\ell\ge 2r_\ell$ for all $\ell\in\mathcal S$ and Assumptions~\ref{ass:A}--\ref{ass:C} hold from iteration $t_0$ onward. Over $T$ steps $t=t_0,\ldots,t_0+T-1$ of ZO-RGD (Algorithm~\ref{alg:zo_rgd}) with constant stepsize
\begin{equation}
    \eta=\min\left\{\frac{1}{2L_R(\bar D+2)},\;
    \sqrt{\frac{\Delta_0}{L_R(\bar D+2)\sigma_D^2T}}\right\},
    \label{eq:formal_stepsize}
\end{equation}
where the second term is taken as $+\infty$ when $\sigma_D=0$, the iterates satisfy
\begin{align}
    &\frac1T\sum_{t=t_0}^{t_0+T-1}\mathbb E\bigl[\|\operatorname{grad}f(\boldsymbol x_t)\|_{\boldsymbol D}^2\bigr]
    \nonumber\\
    &\quad\le
    8\sqrt{\frac{\Delta_0L_R(\bar D+2)\sigma_D^2}{T}}
    +\frac{8\Delta_0L_R(\bar D+2)}{T}
    \nonumber\\
    &\qquad+\frac{\rho_3^2\epsilon^4}{9}\left[(\bar D+2)^4+\frac{(\bar D+6)^4}{2(\bar D+2)}\right].
    \label{eq:formal_rate}
\end{align}
\end{theorem}

\begin{proof}
See the \hyperref[app:theory]{Appendix}.
\end{proof}

In~\eqref{eq:formal_rate}, the first two terms decay at $\mathcal O(T^{-1/2})$ (minibatch noise) and $\mathcal O(T^{-1})$, while the $T$-independent $\mathcal O(\epsilon^4)$ floor reflects finite-difference bias and $L_R$ captures retraction curvature. Since~\eqref{eq:formal_rate} uses the $c_\ell$-weighted norm~\eqref{eq:weighted_norm_main} and ambient dimension $\bar D$, Corollary~\ref{cor:dimension} converts it into an unweighted block-gradient bound whose leading term scales with the manifold dimension.

\begin{corollary}[Manifold-dimension scaling]
\label{cor:dimension}
Let $B_T$ denote the right-hand side of~\eqref{eq:formal_rate}, and define $c_{\min}\triangleq\min_{\ell\in\mathcal S}c_\ell$, $\kappa\triangleq\max_{\ell\in\mathcal S}c_\ell/c_{\min}$, $d_{\mathcal M}\triangleq\sum_{\ell=1}^N d_{r_\ell}$, and $\bar\sigma^2\triangleq\sum_{\ell=1}^N\sigma_\ell^2$. Under the conditions of Theorem~\ref{thm:convergence},
\begin{equation}
    \frac1T\sum_{t=t_0}^{t_0+T-1}\mathbb E\Bigl[\sum_{\ell=1}^N\bigl\|\operatorname{grad}_{\mathcal M_\ell}f(\boldsymbol x_t)\bigr\|_{\mathrm F}^2\Bigr]
    \le\frac{B_T}{c_{\min}},
    \label{eq:cor_blocks}
\end{equation}
whose leading $\mathcal O(T^{-1/2})$ stochastic term satisfies
\begin{align}
    &\frac{8}{c_{\min}}\sqrt{\frac{\Delta_0L_R(\bar D+2)\sigma_D^2}{T}}
    \nonumber\\
    &\quad\le
    8\sqrt{\frac{\Delta_0L_R\bigl(\kappa d_{\mathcal M}+\frac{d_0+2}{c_{\min}}\bigr)\bigl(\kappa\bar\sigma^2+\frac{\sigma_0^2}{c_{\min}}\bigr)}{T}}.
    \label{eq:cor_leading}
\end{align}
\end{corollary}

\begin{proof}
From~\eqref{eq:weighted_norm_main}, $\|\operatorname{grad}f(\boldsymbol x_t)\|_{\boldsymbol D}^2\ge c_{\min}\sum_{\ell=1}^N\|\operatorname{grad}_{\mathcal M_\ell}f(\boldsymbol x_t)\|_{\mathrm F}^2$, so dividing~\eqref{eq:formal_rate} by $c_{\min}$ gives~\eqref{eq:cor_blocks}. Since $m_\ell n_\ell=c_\ell d_{r_\ell}$, we have $\bar D\le\kappa c_{\min}d_{\mathcal M}+d_0$ and $\sigma_D^2\le\kappa c_{\min}\bar\sigma^2+\sigma_0^2$. Substituting both bounds into the leading term of $B_T/c_{\min}$ yields~\eqref{eq:cor_leading}.
\end{proof}

When $c_\ell\equiv c$ and $d_0=0$ ($\sigma_0=0$),~\eqref{eq:cor_leading} reduces to $8\sqrt{\Delta_0L_R(d_{\mathcal M}+2/c)\bar\sigma^2/T}$, replacing $\sum_{\ell=1}^N m_\ell n_\ell$ with the manifold dimension $d_{\mathcal M}$ (about $640\times$ smaller for a $5120\times5120$ block at $r=4$). Meanwhile, the $\mathcal O(\epsilon^4)$ finite-difference floor in~\eqref{eq:formal_rate} retains its ambient dependence on $\bar D$.

\section{Numerical Experiments}
\label{sec:experiments}

\subsection{Setup}
\label{ssec:exp_protocol}

\emph{Tasks, models, and datasets.}
We evaluate Open Pre-trained Transformer (OPT) language models OPT-1.3B and OPT-13B~\cite{zhang2022opt} on the eleven-task suite of MeZO and LOZO~\cite{malladi2023fine,chen2025lozo}. We further evaluate OPT-30B on five tasks from this suite. The suite covers NLU tasks including sentiment analysis (SST-2~\cite{socher2013recursive}) and eight SuperGLUE benchmarks (RTE, CB, BoolQ, WSC, WiC, MultiRC, COPA, and ReCoRD)~\cite{wang2019superglue}, together with generative question answering and reading comprehension (SQuAD~\cite{rajpurkar2016squad} and DROP~\cite{dua2019drop}). To test architectural transfer, we evaluate Llama-3.1-8B~\cite{grattafiori2024llama3} and Qwen2.5-7B~\cite{qwen2025qwen25} on RTE, BoolQ, SQuAD, and mathematical reasoning (GSM8K~\cite{cobbe2021training}). Each run samples up to $1{,}000$ training, $500$ development, and $1{,}000$ evaluation examples ($100$ development examples on CB and COPA). SQuAD and DROP report F1, GSM8K reports exact match, and all remaining tasks report accuracy.

\emph{Protocol.}
All self-run methods train for $20{,}000$ steps at batch size $16$ in 16-bit floating point (FP16) and evaluate development loss every $4{,}000$ steps. The set $\mathcal S$ covers all attention, feed-forward, and embedding matrices (with tied output heads shared), while $\boldsymbol b$ collects 1D biases and normalization scales. Under the unified protocol, each method selects its learning rate from Table~\ref{tab:hparams} by lowest development loss on seed $0$. For each random seed, we evaluate the checkpoint with the lowest development loss once, and report multi-seed results as mean $\pm$ standard deviation (SD). Default settings are $r=4$ for ZO-RGD and $r=2$ with $\beta=0.9$ for ZO-RGD-S in forward module order (embeddings followed by transformer layers); Section~\ref{ssec:ablation} reports a matched-rank ($r=2$) comparison of both variants. Both variants store $(\boldsymbol U_\ell,\boldsymbol\Sigma_\ell,\boldsymbol V_\ell)$ in FP16, execute QR and SVD retractions in 32-bit floating point (FP32) from $\sigma_{\mathrm{init}}=10^{-6}$, and use two forward passes per step.

\begin{table}[!t]
\centering
\footnotesize
\renewcommand{\arraystretch}{1.05}
\setlength{\tabcolsep}{5pt}
\caption{Hyperparameters of the Unified Protocol}
\label{tab:hparams}
\begin{tabular}{@{}lccc@{}}
\toprule
Method & Rank & Learning-rate grid & $\epsilon$ \\
\midrule
MeZO & -- & $\{10^{-7},10^{-6}\}^{\dagger}$ & $10^{-3}$ \\
LOZO & $1$ & $\{10^{-7},10^{-6}\}^{\dagger}$ & $10^{-3}$ \\
MeZO-LoRA, LOZO+LoRA & $8$ & $\{5\times10^{-5},10^{-4}\}^{\ddagger}$ & $10^{-2}$ \\
ZO-RGD & $4$ & $\{10^{-7},10^{-6}\}^{\dagger}$ & $10^{-3}$ \\
ZO-RGD-S & $2$ & $\{10^{-7},10^{-6}\}^{\dagger}$ & $10^{-3}$ \\
\midrule
LoRA+AdamW & $8$ & $\{10^{-5},5\times10^{-5},10^{-4}\}$ & -- \\
\bottomrule
\end{tabular}\par\vspace{3pt}
\parbox{\columnwidth}{\footnotesize On SQuAD and DROP, grids marked $^{\dagger}$ add $5\times10^{-7}$ and the grid marked $^{\ddagger}$ adds $10^{-5}$. LOZO uses $r=1$ and resamples its factor every $100$ steps following its released script, and LoRA baselines use scaling $\alpha=16$.}
\end{table}

\begin{table*}[t]
\centering
\footnotesize
\renewcommand{\arraystretch}{1.05}
\setlength{\tabcolsep}{5pt}
\caption{OPT-1.3B Comparison on the Eleven-Task Suite Under the Unified Protocol (\%)}
\label{tab:breadth_main}
\begin{tabular}{@{}lcccccccccccc@{}}
\toprule
Method & SST-2 & RTE & CB & BoolQ & WSC & WiC & MultiRC & COPA & ReCoRD & SQuAD & DROP & Avg. \\
\midrule
MeZO & 89.33 & \textbf{65.34} & 69.64 & 64.40 & 60.58 & 58.31 & \textbf{60.90} & 76.00 & 70.30 & 75.04 & 24.85 & 64.97 \\
MeZO-LoRA & 90.02 & 63.54 & 69.64 & 65.00 & \textbf{61.54} & 58.62 & 60.30 & 74.00 & 70.50 & 75.84 & 24.82 & 64.89 \\
LOZO & 87.61 & \textbf{65.34} & 67.86 & 64.70 & \textbf{61.54} & 60.03 & 57.20 & 73.00 & 70.20 & 72.71 & 23.27 & 63.95 \\
LOZO+LoRA & 90.71 & 63.54 & \textbf{71.43} & 65.20 & 58.65 & 58.93 & 60.10 & 74.00 & \textbf{71.90} & 75.84 & 25.55 & 65.08 \\
\midrule
ZO-RGD & 92.43 & \textbf{65.34} & \textbf{71.43} & \textbf{66.90} & 55.77 & 57.84 & 59.40 & \textbf{79.00} & 68.60 & 77.66 & \textbf{26.30} & \textbf{65.52} \\
ZO-RGD-S & \textbf{92.55} & 63.90 & \textbf{71.43} & 65.30 & 59.62 & \textbf{61.60} & 59.60 & 70.00 & 67.70 & \textbf{77.96} & 25.97 & 65.06 \\
\bottomrule
\end{tabular}\par\vspace{3pt}
{\footnotesize Every entry is a single run on seed $1$. Bold marks the column maximum, including ties.}
\end{table*}

\emph{Baselines.}
We compare against full-parameter MeZO~\cite{malladi2023fine}, rank-one full-weight LOZO~\cite{chen2025lozo}, and their factorized LoRA counterparts MeZO-LoRA and LOZO+LoRA. HiZOO~\cite{zhao2025hizoo} and first-order LoRA with AdamW (LoRA+AdamW) serve as reference baselines. With the released HiZOO code at $\eta=10^{-7}$ under our protocol, its Hessian scale remains inactive, reducing the update to MeZO at four forward passes per step. Where available, we also include published zero-shot, in-context learning (ICL), RoZO~\cite{rozo2026}, and guiding-vector (GV) candidate-selection LoRA baselines (MeZO-GV and SubZero-GV~\cite{jin2026steering}).

\vspace{-.2cm}
\subsection{Results on Medium-Sized Language Models}
\label{ssec:broad}

Table~\ref{tab:breadth_main} compares all six forward-only methods on OPT-1.3B under the unified protocol. ZO-RGD and ZO-RGD-S lead outright on six tasks and tie for best on RTE and CB, whereas the Euclidean baselines lead on WSC, MultiRC, and ReCoRD. On SST-2 and SQuAD, ZO-RGD-S attains $92.55\%$ accuracy and $77.96\%$ F1, surpassing the strongest baseline LOZO+LoRA ($90.71\%$ and $75.84\%$). Across all eleven tasks, ZO-RGD achieves the highest mean score of $65.52\%$, followed by LOZO+LoRA ($65.08\%$) and ZO-RGD-S ($65.06\%$).

\vspace{-.2cm}
\subsection{Results on Large Language Models}
\label{ssec:large}

\begin{table*}[t]
\centering
\footnotesize
\renewcommand{\arraystretch}{1.05}
\setlength{\tabcolsep}{4.8pt}
\caption{OPT-13B Comparison on the Eleven-Task Suite With Published Baselines (\%)}
\label{tab:main_13b}
\begin{tabular}{@{}lcccccccccccc@{}}
\toprule
Method & SST-2 & RTE & CB & BoolQ & WSC & WiC & MultiRC & COPA & ReCoRD & SQuAD & DROP & Avg. \\
\midrule
Zero-shot & 58.8 & 59.6 & 46.4 & 59.0 & 38.5 & 55.0 & 46.9 & 80.0 & 81.0 & 46.2 & 14.6 & 50.50 \\
ICL & 87.0 & 62.1 & 57.1 & 66.9 & 39.4 & 50.5 & 53.1 & 87.0 & 82.3 & 75.9 & 29.5 & 60.85 \\
\midrule
MeZO & 91.3 & 68.2 & 66.1 & 68.1 & 61.5 & 59.4 & 59.4 & 88.0 & 81.3 & 81.8 & 31.3 & 67.51 \\
MeZO-LoRA & 89.6 & 67.9 & 67.8 & 73.5 & 63.5 & 60.2 & 61.3 & 84.1 & 81.5 & 82.1 & 31.3 & 68.13 \\
LOZO & 91.7 & 70.4 & 69.6 & 71.9 & 63.5 & 60.8 & 63.0 & 89.0 & 81.3 & 84.9 & 30.7 & 69.55 \\
\midrule
RoZO$^{\ddagger}$ & 92.4 & 71.1 & 69.8 & 72.5 & 63.8 & 61.3 & 63.4 & \underline{89.6} & \textbf{84.4} & 85.1 & \textbf{32.7} & 70.17 \\
MeZO-GV (LoRA)$^{\dagger}$ & 91.6 & \underline{72.6} & \underline{72.8} & \underline{75.6} & \textbf{66.3} & 60.9 & 61.9 & 89.0 & 82.9 & 84.9 & \textbf{32.7} & \underline{70.83} \\
SubZero-GV (LoRA)$^{\dagger}$ & \textbf{94.0} & \textbf{75.8} & \textbf{73.8} & \textbf{77.6} & \underline{65.4} & \textbf{63.9} & \textbf{64.1} & \textbf{90.0} & \underline{83.8} & \underline{85.3} & \underline{32.4} & \textbf{72.23} \\
\midrule
ZO-RGD & 92.43 & 72.20 & 66.07 & 71.40 & 59.62$^{a}$ & \underline{62.54} & 57.80 & 87.00 & Fail & \textbf{87.19} & 30.02$^{b}$ & 68.63 \\
ZO-RGD-S & \underline{92.55} & 70.04$^{a}$ & 71.43 & 74.20 & 61.54 & 61.13 & \underline{63.70} & 85.00 & 82.10 & 84.06 & 28.03 & 69.17 \\
\bottomrule
\end{tabular}\par\vspace{3pt}
\parbox{0.94\textwidth}{\footnotesize Zero-shot, ICL, MeZO, MeZO-LoRA, and LOZO are cited from~\cite{chen2025lozo}. $^{\ddagger}$Cited from~\cite{rozo2026}, combining factor-tangent perturbations with momentum, preconditioning, and trust-region scaling over $r\in\{2,\ldots,32\}$ and $q\in\{1,2,4,8\}$. $^{\dagger}$Cited from~\cite{jin2026steering}, using four forward passes per step over $10{,}000$ steps ($40{,}000$ total passes). Each ZO-RGD and ZO-RGD-S entry is a single run selected by its development metric within that variant's sweep over $r\in\{2,4\}$, $\eta\in\{5\times10^{-8},10^{-7},5\times10^{-7}\}$, and $\epsilon\in\{10^{-3},10^{-4}\}$. $^{a}$Uses $\epsilon=10^{-4}$. $^{b}$Caps singular values via $\min\{\sigma_i,8\}$ after retraction. Avg.\ averages the ten tasks excluding ReCoRD. Bold and underline denote the best and second-best column entries, and Fail denotes divergence.}
\end{table*}

Table~\ref{tab:main_13b} compares ZO-RGD and ZO-RGD-S, with hyperparameters selected on the development set, against published OPT-13B baselines. Compared with the published Euclidean two-pass ZO baselines, ZO-RGD-S scores higher on $6$ of the $11$ tasks and ZO-RGD scores higher on $4$ of its $10$ completed tasks. The Euclidean baselines remain higher on WSC, COPA, and DROP. On SQuAD and RTE, ZO-RGD reaches $87.19\%$ F1 and $72.20\%$ accuracy, exceeding LOZO by $2.29$ and $1.80$ points. Across the ten completed tasks, ZO-RGD-S ($69.17\%$) and ZO-RGD ($68.63\%$) match LOZO ($69.55\%$) within one point. Published RoZO~\cite{rozo2026} ($70.17\%$) and the four-pass candidate-selection methods MeZO-GV and SubZero-GV~\cite{jin2026steering} ($70.83\%$ and $72.23\%$) reach higher ten-task averages by incorporating momentum, preconditioning, and trust-region heuristics or multi-candidate perturbation screening, whereas ZO-RGD and ZO-RGD-S isolate the base two-pass manifold update without auxiliary optimizer states.

\vspace{-.2cm}
\subsection{Controlled Comparison}
\label{ssec:controlled_evidence}

\begin{table*}[t]
\centering
\footnotesize
\renewcommand{\arraystretch}{1.05}
\setlength{\tabcolsep}{5.5pt}
\caption{Controlled Comparison Under the Unified Protocol on OPT-13B and OPT-30B (\%)}
\label{tab:controlled_13b}
\begin{tabular}{@{}lccccccccc@{}}
\toprule
& & \multicolumn{2}{c}{OPT-13B (mean $\pm$ SD, 3 seeds)} & \multicolumn{6}{c}{OPT-30B (1 seed)} \\
\cmidrule(lr){3-4}\cmidrule(l){5-10}
Method & Passes/step & RTE & SQuAD & SST-2 & RTE & BoolQ & WSC & WiC & Avg. \\
\midrule
MeZO & 2F & $60.4\pm1.5$ & $82.8\pm1.0$ & 90.37 & 64.98$^{\ast}$ & 66.80 & 61.54 & 50.78 & 66.89 \\
HiZOO$^{\dagger}$ & 4F & $62.1\pm2.6$ & -- & -- & -- & -- & -- & -- & -- \\
LOZO & 2F & $62.1\pm0.6$ & $82.2\pm0.7$ & 90.02 & 63.18 & 69.40 & \textbf{63.46} & \textbf{57.84} & 68.78 \\
\midrule
ZO-RGD ($r=4$) & 2F & $61.9\pm2.9$ & $83.3\pm1.0$ & 92.66 & 67.51 & \textbf{72.70} & 48.08 & 51.25 & 66.44 \\
ZO-RGD-S ($r=2$) & 2F & $\mathbf{69.4\pm3.3}$ & $\mathbf{84.9\pm1.3}$ & \textbf{92.78} & \textbf{72.92}$^{\ast}$ & 72.50 & 58.65 & 50.63 & \textbf{69.50} \\
\midrule
LoRA+AdamW$^{\ddagger}$ & 1F+1B & $83.4\pm2.9$ & OOM & -- & -- & -- & -- & -- & -- \\
\bottomrule
\end{tabular}\par\vspace{3pt}
\parbox{0.92\textwidth}{\footnotesize OPT-13B reports RTE accuracy and SQuAD F1 over seeds $\{1,2,3\}$, while OPT-30B reports single-run accuracies on seed $0$. F and B denote forward and backward passes. Bold marks the best forward-only result in each column, dashes indicate unrun settings, and OOM denotes out-of-memory. $^{\ast}$One of the two grid learning rates completed. $^{\dagger}$Evaluated at $\eta=10^{-7}$ on seeds $\{0,1,2\}$. $^{\ddagger}$First-order reference excluded from forward-only ranking.}
\end{table*}

Table~\ref{tab:controlled_13b} compares MeZO, LOZO, ZO-RGD, and ZO-RGD-S under identical tuning budgets and checkpoint rules. On OPT-13B across three seeds, ZO-RGD-S ($\eta=10^{-6}$ on RTE and $10^{-7}$ on SQuAD) attains $69.4\pm3.3\%$ on RTE and $84.9\pm1.3\%$ F1 on SQuAD. These scores surpass both LOZO ($62.1\pm0.6\%$ and $82.2\pm0.7\%$) and MeZO ($60.4\pm1.5\%$ and $82.8\pm1.0\%$), while ZO-RGD ($83.3\pm1.0\%$) also leads both baselines on SQuAD. First-order LoRA+AdamW reaches $83.4\pm2.9\%$ on RTE using $126.2$\,GB on an H200 GPU. At batch size $16$ without activation checkpointing, it runs out of memory on SQuAD on an $80$\,GB A100 GPU.

Under the same protocol on OPT-30B (Table~\ref{tab:controlled_13b}), ZO-RGD and ZO-RGD-S outperform MeZO and LOZO on SST-2, RTE, and BoolQ, reaching $72.70\%$ and $72.50\%$ versus $69.40\%$ for LOZO on BoolQ. Conversely, LOZO leads on WSC ($63.46\%$) and WiC ($57.84\%$). On those two tasks, $\eta=10^{-6}$ was selected over $\eta=10^{-7}$ by small differences in development loss ($0.003$--$0.014$). At $\eta=10^{-7}$, ZO-RGD and ZO-RGD-S score $61.5\%$ and $63.5\%$ on WSC and $57.2\%$ and $56.9\%$ on WiC. Across all five OPT-30B tasks, ZO-RGD-S achieves the highest mean of $69.50\%$, compared with $68.78\%$ for LOZO, $66.89\%$ for MeZO, and $66.44\%$ for ZO-RGD. SQuAD is omitted on OPT-30B since a single forward pass exceeds $80$\,GB on an A100 GPU.

\vspace{-.2cm}
\subsection{Convergence Under a Shared Learning Rate}
\label{ssec:convergence}

\begin{figure}[t]
\centering
\includegraphics[width=\columnwidth]{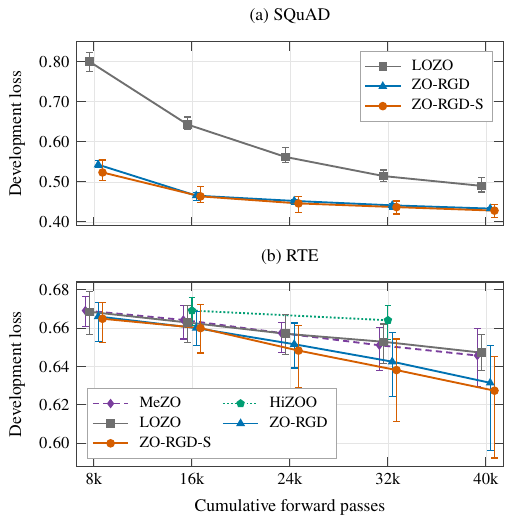}
\caption{Development loss on OPT-13B versus cumulative forward passes at the shared learning rate $\eta=10^{-7}$ for (a) SQuAD and (b) RTE. Curves plot sample means and whiskers span the minimum-to-maximum range over finite evaluations at each checkpoint, with markers slightly offset horizontally. LOZO, ZO-RGD, and ZO-RGD-S launch five seeds (on SQuAD, three ZO-RGD seeds remain finite through $32{,}000$ passes and two at $40{,}000$), while MeZO and HiZOO use three seeds on RTE. HiZOO uses four forward passes per step (yielding checkpoints at $16{,}000$ and $32{,}000$ passes within the $40{,}000$-pass window), whereas all other methods use two.}
\label{fig:convergence_13b}
\end{figure}

Fig.~\ref{fig:convergence_13b} plots held-out development loss on OPT-13B against cumulative forward passes at the shared learning rate $\eta=10^{-7}$ over finite evaluations. In Fig.~\ref{fig:convergence_13b}(a) on SQuAD, ZO-RGD and ZO-RGD-S descend faster than LOZO from the first checkpoint and stay below LOZO's min--max band throughout training. At $40{,}000$ passes, they reach development losses of $0.433$ and $0.428$, compared with $0.490$ for LOZO. Their best-checkpoint test F1 scores reach $85.2\pm1.0\%$ (across the three finite runs) and $85.0\pm0.9\%$ (across five runs), versus $81.8\pm0.9\%$ for LOZO. This descent reflects both tangent-space alignment and the blockwise scaling $c_\ell$ in Proposition~\ref{prop:tangent_sphere}. In Fig.~\ref{fig:convergence_13b}(b) on RTE, the mean loss curves of ZO-RGD and ZO-RGD-S drop below MeZO and LOZO after $16{,}000$ passes. Their seed ranges still overlap, and final RTE accuracies remain within one SD. Consequently, the SQuAD improvement holds under both tuned and shared learning rates, whereas the RTE gain of ZO-RGD-S in Table~\ref{tab:controlled_13b} comes from selecting $\eta=10^{-6}$. At $\eta=10^{-7}$ on SQuAD, ZO-RGD ($r=4$) diverges before step $4{,}000$ on two of five seeds (steps $500$ and $1{,}500$) and after step $16{,}000$ on a third, whereas ZO-RGD-S ($r=2$) completes all five seeds.

\vspace{-.2cm}
\subsection{Additional Language Model Architectures}
\label{ssec:modern_main}

\begin{table*}[t]
\centering
\footnotesize
\renewcommand{\arraystretch}{1.05}
\setlength{\tabcolsep}{6pt}
\caption{Evaluation on Llama-3.1-8B and Qwen2.5-7B (Mean $\pm$ SD Over Three Seeds, \%)}
\label{tab:modern_panel}
\begin{tabular}{@{}lcccccccc@{}}
\toprule
& \multicolumn{4}{c}{Llama-3.1-8B} & \multicolumn{4}{c}{Qwen2.5-7B} \\
\cmidrule(lr){2-5}\cmidrule(lr){6-9}
Method & RTE & BoolQ & SQuAD & GSM8K & RTE & BoolQ & SQuAD & GSM8K \\
\midrule
LOZO & $56.9\pm7.1$ & $61.6\pm2.7$ & $86.8\pm2.2$ & $31.2\pm2.4$ & $86.6\pm1.9$ & $80.7\pm0.6$ & $87.0\pm0.8$ & $\mathbf{74.2\pm2.8}$ \\
ZO-RGD & $62.5\pm3.2$ & $66.9\pm1.5$ & $\mathbf{87.1\pm0.8}$ & $\mathbf{37.7\pm5.8}$ & $\mathbf{88.3\pm0.8}$ & $\mathbf{82.4\pm1.2}$ & $\mathbf{87.9\pm0.3}$ & $73.2\pm2.8$ \\
ZO-RGD-S & $\mathbf{62.9\pm8.4}$ & $\mathbf{85.8\pm1.1}$ & $85.5\pm1.0$ & $34.0\pm1.0$ & $\mathbf{88.3\pm1.0}$ & $82.3\pm0.5$ & $87.6\pm0.2$ & $70.3\pm2.8$ \\
\bottomrule
\end{tabular}\par\vspace{3pt}
{\footnotesize RTE and BoolQ report accuracy, SQuAD reports F1, and GSM8K reports exact match on $n=200$ (sampling standard error $3$--$3.5$ points) over seeds $\{0,1,2\}$. Bold marks the column maximum, including ties on mean score.}
\end{table*}

Table~\ref{tab:modern_panel} evaluates transfer to Llama-3.1-8B and Qwen2.5-7B across three seeds, with learning rates selected by a $2{,}000$-step pilot on a disjoint seed. On Qwen2.5-7B, both proposed methods outperform LOZO on RTE, BoolQ, and SQuAD, while LOZO leads on GSM8K. On Llama-3.1-8B, ZO-RGD surpasses LOZO across all four benchmarks, including $62.5\pm3.2\%$ versus $56.9\pm7.1\%$ on RTE. The wide BoolQ gap on Llama-3.1-8B stems from the pilot rule selecting $\eta=10^{-7}$ for ZO-RGD-S and $10^{-6}$ for ZO-RGD and LOZO. At a shared $\eta=10^{-7}$ on seed $0$, ZO-RGD, ZO-RGD-S, and LOZO achieve $85.5\%$, $85.5\%$, and $80.9\%$, respectively.

\vspace{-.2cm}
\subsection{Ablation Studies}
\label{ssec:ablation}
\label{ssec:geometry_main}

\begin{table}[!t]
\centering
\footnotesize
\renewcommand{\arraystretch}{1.05}
\setlength{\tabcolsep}{4pt}
\caption{OPT-1.3B Controls of the Estimator (Mean $\pm$ SD Over Three Paired Seeds, \%)}
\label{tab:mechanism_1p3b}
\begin{tabular}{@{}lc@{}}
\toprule
(a) Finite-difference rule on SST-2 & Accuracy \\
\midrule
Tangent-projected, normalized ($\widetilde{\boldsymbol Z}_t$, proposed) & $\mathbf{92.4\pm0.2}$ \\
Retracted evaluation points ($\mathcal R_{\Delta\boldsymbol W_t}(\pm\epsilon\widetilde{\boldsymbol Z}_t)$) & $89.6\pm0.9$ \\
Rank-$r$ outer product, learning rate $10^{-7}$ & $54.3\pm1.1$ \\
Rank-$r$ outer product, learning rate $2\times10^{-6}$ & $64.7\pm3.9$ \\
\bottomrule
\end{tabular}

\vspace{4pt}
\begin{tabular}{@{}lcc@{}}
\toprule
(b) Task & Manifold ($\Delta\boldsymbol W\in\mathcal M_4$) & Euclidean $\boldsymbol B\boldsymbol A$ factors \\
\midrule
SST-2 (acc.) & $\mathbf{87.7\pm0.6}$ & $78.5\pm10.0$ \\
RTE (acc.) & $56.0\pm1.4$ & $\mathbf{58.4\pm4.0}$ \\
SQuAD (F1) & $\mathbf{71.3\pm0.5}$ & $67.7\pm1.3$ \\
\bottomrule
\end{tabular}\par\vspace{3pt}
\parbox{\columnwidth}{\footnotesize Both subtables train for $20{,}000$ steps. Subtable (a) uses $r=4$ and $\eta=10^{-7}$ in rows 1--3 on seeds $\{0,1,2\}$ (seed $1$ matches Table~\ref{tab:breadth_main}). In (b), a $2{,}000$-step pilot selects $\eta=10^{-6},10^{-7},10^{-6}$ on SST-2, RTE, and SQuAD for $\mathcal M_4$ and $10^{-4}$ for $(\boldsymbol B,\boldsymbol A)$ (FP32 factors), shared with 1D parameters on seeds $\{1,2,3\}$.}
\end{table}

\emph{Finite-difference rule and parameterization.}
Table~\ref{tab:mechanism_1p3b}(a) compares four perturbation rules on OPT-1.3B SST-2 using the same truncated-SVD update retraction~\eqref{eq:retraction}. Ambient tangent-line queries $\widetilde{\boldsymbol Z}_t$ achieve $92.4\pm0.2\%$, improving over retracted manifold queries $\mathcal R_{\Delta\boldsymbol W_t}(\pm\epsilon\widetilde{\boldsymbol Z}_t)$ ($89.6\pm0.9\%$) while avoiding two SVDs per step. Under the same fixed-rank SVD retraction, replacing $\widetilde{\boldsymbol Z}_t$ with a Frobenius-normalized rank-$r$ outer product $\boldsymbol P\boldsymbol Q^\top$ drops accuracy to $54.3\pm1.1\%$ at $\eta=10^{-7}$ and $64.7\pm3.9\%$ at $\eta=2\times10^{-6}$. Within this retracted-state control, tangent projection exceeds the outer-product probe by at least $27.7$ points. In Table~\ref{tab:mechanism_1p3b}(b), manifold updates on $\mathcal M_4$ achieve higher accuracy and lower variance than Euclidean ZO on $(\boldsymbol B,\boldsymbol A)$ for SST-2 and SQuAD, while trailing on RTE. On SST-2, this $2{,}000$-step pilot selects $\eta=10^{-6}$ and yields $87.7\pm0.6\%$, compared with $92.4\pm0.2\%$ at $\eta=10^{-7}$ in (a).

\emph{Normalization.}
Without~\eqref{eq:zo_norm}, tangent projection shrinks perturbation entries by $\sqrt{d_{r_\ell}/(m_\ell n_\ell)}\approx0.045$--$0.063$ on OPT-1.3B, approaching the FP16 rounding floor at $\epsilon=10^{-3}$. Removing~\eqref{eq:zo_norm} at $\eta=10^{-7}$ drops SST-2 accuracy to $58.1\%$ on seed $0$ and causes SQuAD runs to diverge. Scaling $\eta$ globally by $\sum_\ell m_\ell n_\ell/\sum_\ell d_{r_\ell}=350$ (shared with 1D parameters) recovers $90.5\pm0.2\%$ versus $92.7\pm0.7\%$ on SST-2, while two of three compensated SQuAD seeds still diverge ($73.3\%$ F1 on the completed seed). Thus,~\eqref{eq:zo_norm} provides per-block query and step-size normalization that a global learning-rate multiplier does not reproduce.

\emph{Scalar smoothing and block ordering.}
At matched rank $r=2$ and identical learning rates on OPT-13B, both ZO-RGD-S and ZO-RGD complete all three seeds, scoring $69.43\pm3.28\%$ and $64.50\pm3.26\%$ on RTE ($p=0.28$, paired $t$-test) and $84.89\pm1.30\%$ and $85.38\pm0.34\%$ F1 on SQuAD ($p=0.47$). Neither paired difference is statistically significant across three seeds. On OPT-1.3B SST-2 across three paired seeds, forward, reverse, and shuffled block orderings of ZO-RGD-S attain $91.6\pm0.8\%$, $91.8\pm0.9\%$, and $91.7\pm0.4\%$, respectively.

\emph{Rank and conditioning of the update.}
For LOZO, the accumulated update $\boldsymbol W_T-\boldsymbol W_0$ reaches a median layerwise effective rank ($90\%$ spectral energy) of $1{,}019/2{,}048$ on OPT-1.3B and $2{,}466/5{,}120$ on OPT-13B. By contrast, ZO-RGD maintains exact rank $r$ at every checkpoint. Across $20$ completed instrumented runs on OPT-13B and OPT-30B, $\min_\ell\sigma_{r_\ell}(\Delta\boldsymbol W_{t,\ell})$ exceeds $0.08$ from step $500$ onward and reaches $0.52$--$1.15$ at termination, with condition numbers at most $1.07$.

\vspace{-.2cm}
\subsection{Memory and Wall-Clock Time}
\label{ssec:cost_main}

\begin{table}[!t]
\centering
\footnotesize
\renewcommand{\arraystretch}{1.05}
\setlength{\tabcolsep}{4.5pt}
\caption{Per-Step Runtime and Peak GPU Memory on OPT-13B at Batch Size 16}
\label{tab:cost_13b}
\begin{tabular}{@{}lccc@{}}
\toprule
Method & Passes/step & Time/step (s) & Peak reserved (GB) \\
\midrule
MeZO & 2F & 2.03 & 39.6 \\
LOZO & 2F & 1.93 & 39.4 \\
ZO-RGD & 2F & 3.51 & 39.4 \\
ZO-RGD-S & 2F & 3.54 & 39.4 \\
\midrule
LoRA+AdamW & 1F+1B & OOM (A100) & 126.2 (H200, RTE) \\
\bottomrule
\end{tabular}\par\vspace{3pt}
\parbox{\columnwidth}{\footnotesize F and B denote forward and backward passes. Forward-only methods average two $1{,}000$-step SQuAD runs across two A100-80G GPUs. LoRA+AdamW hits OOM on SQuAD (A100-80G) without activation checkpointing and uses $126.2$\,GB on RTE (H200).}
\end{table}

Table~\ref{tab:cost_13b} reports per-step latency and peak reserved GPU memory on OPT-13B at batch size $16$. On SQuAD, ZO-RGD and ZO-RGD-S match the $39.4$\,GB forward-pass memory footprint of LOZO and MeZO ($39.6$\,GB), while storing all 2D rank-$4$ factors on OPT-1.3B takes $7.5$\,MB in FP16 (plus $1.3$\,MB for 1D parameters). By contrast, LoRA+AdamW exhausts an $80$\,GB A100 GPU on SQuAD and requires $126.2$\,GB on RTE. Per-step latency is $3.51$\,s for ZO-RGD and $3.54$\,s for ZO-RGD-S versus $1.93$\,s for LOZO ($1.8\times$). Generating dense $m_\ell\times n_\ell$ Gaussian matrices from seeds accounts for $48\%$ of step time, and tangent projection accounts for $12\%$.

\vspace{-.2cm}
\subsection{Discussion}
\label{ssec:discussion}

\emph{Evaluation protocols.}
Tables~\ref{tab:breadth_main},~\ref{tab:controlled_13b}, and~\ref{tab:modern_panel} evaluate all methods under matched development-loss checkpoint selection, whereas Table~\ref{tab:main_13b} compares our methods, with hyperparameters selected on the development set, against published results. Self-run baselines in Table~\ref{tab:controlled_13b} score below published numbers, such as $62.1\pm0.6\%$ versus $70.4\%$ for LOZO on OPT-13B RTE, because our unified protocol fixes default script hyperparameters and selects only the learning rate by $4{,}000$-step development loss rather than searching per-task rank, interval, and radius grids~\cite{chen2025lozo}. Moreover, single-seed panels in Table~\ref{tab:breadth_main} and OPT-30B carry no variance estimates. On OPT-1.3B, trained scores on WiC, COPA, and ReCoRD stay close to the zero-shot floor ($57.7\%$, $75.0\%$, and $69.9\%$).

\emph{Limitations.}
Accuracy gains remain task-dependent. Ten-task OPT-13B averages match LOZO within one point and trail multi-component and candidate-selection baselines~\cite{rozo2026,jin2026steering}, while OPT-30B WSC and WiC show small differences in development loss across candidate learning rates. Whenever GPU memory permits backpropagation, first-order LoRA remains substantially higher. On stability and precision, ZO-RGD ($r=4$) diverges early on two of five shared-rate OPT-13B SQuAD seeds, turns non-finite after step $16{,}000$ on a third, and diverges on ReCoRD. Using 16-bit brain floating-point precision trails FP16 by $14.9$ points on OPT-1.3B SST-2 and diverges on OPT-13B SQuAD. Dense Gaussian regeneration also adds a $1.8\times$ per-step latency over LOZO. Finally, Theorem~\ref{thm:convergence} analyzes ZO-RGD on smooth objectives from $t_0$ onward under trajectory regularity, excluding nonsmooth activation kinks in OPT. Meanwhile, formal convergence guarantees for ZO-RGD-S remain open, and a controlled rerun of RoZO~\cite{rozo2026} under our unified protocol remains unavailable in the absence of public code.

\section{Conclusion}
\label{sec:conclusion}

This paper formulates forward-only LLM fine-tuning directly on fixed-rank matrix manifolds. Restricting each additive update to a fixed-rank manifold preserves a compact low-rank adapter throughout training without factor-scaling ambiguity. To offset norm shrinkage after tangent projection, we combined the normalized tangent ZO estimator with truncated-SVD retraction in ZO-RGD and its scalar-smoothed variant ZO-RGD-S. Our theoretical analysis established first-order query equivalence, alignment in expectation with the Riemannian gradient, invariance to factor parameterization, and a nonconvex stationarity bound for ZO-RGD. Experiments on NLU tasks across language models from 1.3 billion to 30 billion parameters showed competitive accuracy against forward-only baselines at inference-level memory cost. Of the two variants, ZO-RGD has no smoothing parameter and is covered by our stationarity bound. ZO-RGD-S with $r=2$ completed all five shared-rate SQuAD runs on OPT-13B, whereas ZO-RGD with $r=4$ diverged on three of them. At matched rank $r=2$, the two variants reached comparable accuracy. Future work includes sampling tangent perturbations directly in factorized coordinates to reduce per-step runtime and integrating candidate-pool directional selection.

\appendix[Convergence Analysis of ZO-RGD]
\label{app:theory}

\setlength{\abovedisplayskip}{4pt plus 1pt minus 1pt}
\setlength{\belowdisplayskip}{4pt plus 1pt minus 1pt}
\setlength{\abovedisplayshortskip}{2pt plus 1pt}
\setlength{\belowdisplayshortskip}{3pt plus 1pt}

This appendix proves Theorem~\ref{thm:convergence} for ZO-RGD (Algorithm~\ref{alg:zo_rgd}) on $\mathcal X$. Lemmas~\ref{lem:probe} and~\ref{lem:bias} bound the moments of the normalized directions and the finite-difference remainder. Lemma~\ref{lem:first_moment} combines them into the estimator moments, and Lemma~\ref{lem:retraction} establishes a one-step descent inequality.

\subsection*{Technical Lemmas}

At a fixed iteration $t$ (omitting subscript $t$ on block quantities), write $d_\ell\triangleq d_{r_\ell}=r_\ell(m_\ell+n_\ell-r_\ell)$, $\rho_\ell\triangleq\sqrt{m_\ell n_\ell}$, $T_\ell\triangleq T_{\Delta\boldsymbol W_\ell}\mathcal M_\ell$, and $P_\ell\triangleq P_{T_\ell}$ for each $\ell\in\mathcal S$, so $c_\ell=\rho_\ell^2/d_\ell\ge1$. For any $\boldsymbol v=(\boldsymbol v_1,\ldots,\boldsymbol v_N,\boldsymbol v_0)\in T_{\boldsymbol x}\mathcal X=\bigl(\prod_{\ell=1}^N T_\ell\bigr)\times\mathbb R^{d_0}$, the operator $\boldsymbol D$ in Section~\ref{ssec:stationarity} gives
\begin{equation}
    \|\boldsymbol v\|_{\boldsymbol D}^2
    =\sum_{\ell=1}^N c_\ell\|\boldsymbol v_\ell\|_{\mathrm F}^2+\|\boldsymbol v_0\|^2
    \ge\|\boldsymbol v\|^2.
    \label{eq:app_weighted_norm}
\end{equation}

\begin{lemma}[Random-direction moments]
\label{lem:probe}
Conditioned on $\mathcal F_t$, the normalized directions $\boldsymbol u_\ell\triangleq\widehat{\boldsymbol Z}_\ell/\|\widehat{\boldsymbol Z}_\ell\|_{\mathrm F}$ are mutually independent and uniformly distributed on the unit spheres of $T_\ell$, with $\|\widetilde{\boldsymbol Z}_\ell\|_{\mathrm F}=\rho_\ell$. For any $\mathcal F_t$-measurable $\boldsymbol g\in T_\ell$, the second-moment identities on $T_\ell$ are given by
\begin{align}
    \mathbb E\bigl[\langle\boldsymbol g,\widetilde{\boldsymbol Z}_\ell\rangle_{\mathrm F}\widetilde{\boldsymbol Z}_\ell\mid\mathcal F_t\bigr]
    &=c_\ell\boldsymbol g,
    \label{eq:probe_moments_2d_vec}\\
    \mathbb E\bigl[\langle\boldsymbol g,\widetilde{\boldsymbol Z}_\ell\rangle_{\mathrm F}^2\mid\mathcal F_t\bigr]
    &=c_\ell\|\boldsymbol g\|_{\mathrm F}^2.
    \label{eq:probe_moments_2d_scal}
\end{align}
Similarly, for any $\mathcal F_t$-measurable vector $\boldsymbol h\in\mathbb R^{d_0}$, the standard Gaussian perturbation $\boldsymbol z_0$ satisfies
\begin{align}
    \mathbb E\bigl[\langle\boldsymbol h,\boldsymbol z_0\rangle\boldsymbol z_0\mid\mathcal F_t\bigr]
    &=\boldsymbol h,
    \label{eq:probe_moments_1d_vec}\\
    \mathbb E\bigl[\langle\boldsymbol h,\boldsymbol z_0\rangle^2\|\boldsymbol z_0\|^2\mid\mathcal F_t\bigr]
    &=(d_0+2)\|\boldsymbol h\|^2.
    \label{eq:probe_moments_1d_scal}
\end{align}
Moreover, with $Q\triangleq\sum_{\ell=1}^N\rho_\ell^2$, the squared norm $\|\boldsymbol\zeta_t\|^2=Q+\|\boldsymbol z_0\|^2$ satisfies
\begin{equation}
    \mathbb E\bigl[\|\boldsymbol\zeta_t\|^4\mid\mathcal F_t\bigr]\le(\bar D+2)^2,
    \quad
    \mathbb E\bigl[\|\boldsymbol\zeta_t\|^8\mid\mathcal F_t\bigr]\le(\bar D+6)^4.
    \label{eq:u_moments}
\end{equation}
\end{lemma}

\begin{proof}
Orthogonal projection of an isotropic Gaussian onto $T_\ell$ yields an isotropic Gaussian on $T_\ell$, whose normalization $\boldsymbol u_\ell$ is uniform on the unit sphere of $T_\ell$ with covariance $(1/d_\ell)\boldsymbol I_{T_\ell}$. Multiplying by $\rho_\ell^2=c_\ell d_\ell$ yields~\eqref{eq:probe_moments_2d_vec}--\eqref{eq:probe_moments_2d_scal}. Standard Gaussian fourth-moment identities give~\eqref{eq:probe_moments_1d_vec}--\eqref{eq:probe_moments_1d_scal}. Since $\|\boldsymbol z_0\|^2\sim\chi^2_{d_0}$ satisfies $\mathbb E[\|\boldsymbol z_0\|^{2j}]=d_0(d_0+2)\cdots(d_0+2j-2)\le(d_0+2k-2)^j$ for $j\le k$, binomial expansion of $\mathbb E[(Q+\|\boldsymbol z_0\|^2)^k]$ yields $\mathbb E[\|\boldsymbol\zeta_t\|^{2k}]\le(Q+d_0+2k-2)^k=(\bar D+2k-2)^k$ for $k\in\{2,4\}$.
\end{proof}

\begin{lemma}[Symmetric finite-difference bias]
\label{lem:bias}
Under Assumption~\ref{ass:B}, with $s^{\mathrm{lin}}_t\triangleq\langle\nabla F(\boldsymbol x_t;\xi_t),\boldsymbol\zeta_t\rangle$, we have
\begin{equation}
    \bigl|\hat s_t-s^{\mathrm{lin}}_t\bigr|\le\frac{\rho_3}{6}\,\epsilon^2\,\|\boldsymbol\zeta_t\|^3.
    \label{eq:two_point_bias}
\end{equation}
\end{lemma}

\begin{proof}
Expanding $\phi(\tau)=F(\boldsymbol x_t+\tau\boldsymbol\zeta_t;\xi_t)$ to third order with Lagrange remainder at $\tau=\pm\epsilon$ gives $\phi(\pm\epsilon)=\phi(0)\pm\epsilon\phi'(0)+\frac{\epsilon^2}{2}\phi''(0)\pm\frac{\epsilon^3}{6}\phi'''(\tau_\pm)$ for $\tau_\pm\in(-\epsilon,\epsilon)$. Subtracting the two expansions cancels $\phi(0)$ and $\phi''(0)$, and applying $|\phi'''(\tau_\pm)|\le\rho_3\|\boldsymbol\zeta_t\|^3$ gives~\eqref{eq:two_point_bias}.
\end{proof}

\begin{lemma}[First and second moments of the estimator]
\label{lem:first_moment}
\label{lem:second_moment}
Under Assumptions~\ref{ass:A}--\ref{ass:C}, the conditional mean of $\widehat{\boldsymbol G}_t=\hat s_t\boldsymbol\zeta_t$ satisfies
\begin{equation}
    \mathbb E\bigl[\widehat{\boldsymbol G}_t\mid\mathcal F_t\bigr]
    =
    \boldsymbol D\operatorname{grad}f(\boldsymbol x_t)+\boldsymbol\beta_t,
    \label{eq:first_moment}
\end{equation}
where the bias vector $\boldsymbol\beta_t\in T_{\boldsymbol x_t}\mathcal X$ is bounded by $\|\boldsymbol\beta_t\|\le B_\epsilon\triangleq(\rho_3/6)\epsilon^2(\bar D+2)^2$. Furthermore, the conditional second moment of $\widehat{\boldsymbol G}_t$ satisfies
\begin{align}
    \mathbb E\bigl[\|\widehat{\boldsymbol G}_t\|^2\mid\mathcal F_t\bigr]
    &\le
    2(\bar D+2)\bigl(\|\operatorname{grad}f(\boldsymbol x_t)\|_{\boldsymbol D}^2+\sigma_D^2\bigr)
    \nonumber\\
    &\quad+\frac{\rho_3^2}{18}\epsilon^4(\bar D+6)^4.
    \label{eq:second_moment}
\end{align}
\end{lemma}

\begin{proof}
We split $\hat s_t$ into its linear part and the finite-difference remainder, and apply the moment identities of Lemma~\ref{lem:probe} to each part. By Lemma~\ref{lem:bias}, we have $\hat s_t=s^{\mathrm{lin}}_t+\delta_t$ with $|\delta_t|\le(\rho_3/6)\epsilon^2\|\boldsymbol\zeta_t\|^3$. Since $\widetilde{\boldsymbol Z}_\ell\in T_\ell$, the linear part is
\begin{equation}
    s^{\mathrm{lin}}_t
    =\sum_{\ell=1}^N\bigl\langle P_\ell\nabla_\ell F(\boldsymbol x_t;\xi_t),\widetilde{\boldsymbol Z}_\ell\bigr\rangle_{\mathrm F}
    +\bigl\langle\nabla_0 F(\boldsymbol x_t;\xi_t),\boldsymbol z_0\bigr\rangle.
    \label{eq:slin_blocks}
\end{equation}
The block directions are independent with zero mean, so the cross-block terms vanish in expectation, and~\eqref{eq:probe_moments_2d_vec} and~\eqref{eq:probe_moments_1d_vec} give
\begin{align}
    \mathbb E\bigl[s^{\mathrm{lin}}_t\boldsymbol\zeta_t\mid\mathcal F_t,\xi_t\bigr]
    &=\bigl(c_1P_1\nabla_1F,\ldots,c_NP_N\nabla_NF,\nabla_0F\bigr)
    \nonumber\\
    &=\boldsymbol D\operatorname{grad}F(\boldsymbol x_t;\xi_t).
\end{align}
Taking the expectation over $\xi_t$ under Assumption~\ref{ass:C} gives $\boldsymbol D\operatorname{grad}f(\boldsymbol x_t)$. The remaining term $\boldsymbol\beta_t=\mathbb E[\delta_t\boldsymbol\zeta_t\mid\mathcal F_t]$ satisfies
\begin{equation}
    \|\boldsymbol\beta_t\|
    \le\frac{\rho_3}{6}\epsilon^2\,\mathbb E\bigl[\|\boldsymbol\zeta_t\|^4\mid\mathcal F_t\bigr]
    \le\frac{\rho_3}{6}\epsilon^2(\bar D+2)^2
\end{equation}
by~\eqref{eq:u_moments}, which proves~\eqref{eq:first_moment}. For the second moment, $\hat s_t^2\le2(s^{\mathrm{lin}}_t)^2+2\delta_t^2$ gives
\begin{equation}
    \|\widehat{\boldsymbol G}_t\|^2
    \le2(s^{\mathrm{lin}}_t)^2\|\boldsymbol\zeta_t\|^2+2\delta_t^2\|\boldsymbol\zeta_t\|^2.
    \label{eq:G_split}
\end{equation}
Since $\|\boldsymbol\zeta_t\|^2=Q+\|\boldsymbol z_0\|^2$ with $Q$ constant,~\eqref{eq:probe_moments_2d_scal} and~\eqref{eq:probe_moments_1d_scal} bound the first term of~\eqref{eq:G_split} as
\begin{align}
    \mathbb E\bigl[(s^{\mathrm{lin}}_t)^2\|\boldsymbol\zeta_t\|^2\mid\mathcal F_t,\xi_t\bigr]
    &\le(Q+d_0+2)\|\operatorname{grad}F(\boldsymbol x_t;\xi_t)\|_{\boldsymbol D}^2
    \nonumber\\
    &=(\bar D+2)\|\operatorname{grad}F(\boldsymbol x_t;\xi_t)\|_{\boldsymbol D}^2,
\end{align}
and Assumption~\ref{ass:C} gives $\mathbb E_{\xi_t}\|\operatorname{grad}F(\boldsymbol x_t;\xi_t)\|_{\boldsymbol D}^2\le\|\operatorname{grad}f(\boldsymbol x_t)\|_{\boldsymbol D}^2+\sigma_D^2$. The second term of~\eqref{eq:G_split} satisfies $\mathbb E[\delta_t^2\|\boldsymbol\zeta_t\|^2\mid\mathcal F_t]\le(\rho_3^2/36)\epsilon^4\,\mathbb E[\|\boldsymbol\zeta_t\|^8\mid\mathcal F_t]\le(\rho_3^2/36)\epsilon^4(\bar D+6)^4$ by~\eqref{eq:u_moments}. Combining the two bounds yields~\eqref{eq:second_moment}.
\end{proof}

\begin{lemma}[Truncated-SVD retraction error and descent inequality]
\label{lem:retraction}
\label{lem:descent}
Let $\boldsymbol X\in\mathcal M_\ell$ with $\sigma_{r_\ell}(\boldsymbol X)\ge\sigma_{\min}>0$ and $\boldsymbol Y\in T_{\boldsymbol X}\mathcal M_\ell$, where $m_\ell,n_\ell\ge 2r_\ell$. Then the truncated-SVD retraction satisfies
\begin{equation}
    \bigl\|\mathcal H_{r_\ell}(\boldsymbol X+\boldsymbol Y)-(\boldsymbol X+\boldsymbol Y)\bigr\|_{\mathrm F}
    \le
    \frac{2\|\boldsymbol Y\|_{\mathrm F}^2}{\sigma_{\min}}.
    \label{eq:retraction_error}
\end{equation}
Consequently, under Assumptions~\ref{ass:A}--\ref{ass:C} with $L_R$ in~\eqref{eq:lr_sigmad}, every iterate of ZO-RGD with $t\ge t_0$ satisfies almost surely
\begin{equation}
    f(\boldsymbol x_{t+1})
    \le
    f(\boldsymbol x_t)
    -\eta\bigl\langle\operatorname{grad}f(\boldsymbol x_t),\widehat{\boldsymbol G}_t\bigr\rangle
    +\frac{L_R}{2}\eta^2\|\widehat{\boldsymbol G}_t\|^2.
    \label{eq:descent}
\end{equation}
\end{lemma}

\begin{proof}
We bound the distance from $\boldsymbol X+\boldsymbol Y$ to $\mathcal H_{r_\ell}(\boldsymbol X+\boldsymbol Y)$ across large and small tangent steps, and insert that bound into the smoothness inequality of $f$. When $\|\boldsymbol Y\|_{\mathrm F}>\sigma_{r_\ell}(\boldsymbol X)/2\ge\sigma_{\min}/2$, $\boldsymbol X\in\mathcal M_\ell$ has rank $r_\ell$ and $\mathcal H_{r_\ell}(\boldsymbol X+\boldsymbol Y)$ minimizes the Frobenius distance to $\boldsymbol X+\boldsymbol Y$ over all matrices of rank at most $r_\ell$. Hence $\|\mathcal H_{r_\ell}(\boldsymbol X+\boldsymbol Y)-(\boldsymbol X+\boldsymbol Y)\|_{\mathrm F}\le\|\boldsymbol X-(\boldsymbol X+\boldsymbol Y)\|_{\mathrm F}=\|\boldsymbol Y\|_{\mathrm F}<2\|\boldsymbol Y\|_{\mathrm F}^2/\sigma_{\min}$. Suppose now that $\|\boldsymbol Y\|_{\mathrm F}\le\sigma_{r_\ell}(\boldsymbol X)/2$, and let $\boldsymbol X=\boldsymbol U\boldsymbol\Sigma\boldsymbol V^\top$ be a thin SVD. By~\eqref{eq:tangent_space}, we write $\boldsymbol Y=\boldsymbol U\boldsymbol M\boldsymbol V^\top+\boldsymbol U_p\boldsymbol V^\top+\boldsymbol U\boldsymbol V_p^\top$ with $\boldsymbol U^\top\boldsymbol U_p=\boldsymbol 0$ and $\boldsymbol V^\top\boldsymbol V_p=\boldsymbol 0$. Since $m_\ell,n_\ell\ge2r_\ell$, we choose $\boldsymbol Q_U\in\mathbb R^{m_\ell\times r_\ell}$ and $\boldsymbol Q_V\in\mathbb R^{n_\ell\times r_\ell}$ with orthonormal columns orthogonal to $\boldsymbol U$ and $\boldsymbol V$, respectively, such that $\boldsymbol U_p=\boldsymbol Q_U\boldsymbol R_U$ and $\boldsymbol V_p=\boldsymbol Q_V\boldsymbol R_V$, giving
\begin{equation}
    \boldsymbol X+\boldsymbol Y
    =\begin{bmatrix}\boldsymbol U&\boldsymbol Q_U\end{bmatrix}
    \begin{bmatrix}\boldsymbol\Sigma+\boldsymbol M&\boldsymbol R_V^\top\\ \boldsymbol R_U&\boldsymbol 0\end{bmatrix}
    \begin{bmatrix}\boldsymbol V&\boldsymbol Q_V\end{bmatrix}^\top,
\end{equation}
where the middle factor is denoted by $\boldsymbol K\in\mathbb R^{2r_\ell\times2r_\ell}$.
The two outer factors have orthonormal columns, so $\|\boldsymbol X+\boldsymbol Y-\mathcal H_{r_\ell}(\boldsymbol X+\boldsymbol Y)\|_{\mathrm F}=\|\boldsymbol K-\mathcal H_{r_\ell}(\boldsymbol K)\|_{\mathrm F}$ and $\|\boldsymbol M\|_{\mathrm F}^2+\|\boldsymbol R_U\|_{\mathrm F}^2+\|\boldsymbol R_V\|_{\mathrm F}^2=\|\boldsymbol Y\|_{\mathrm F}^2$. Applying Weyl's singular-value inequality $\sigma_i(\boldsymbol A+\boldsymbol E)\ge\sigma_i(\boldsymbol A)-\|\boldsymbol E\|_2$ with $\boldsymbol A=\boldsymbol\Sigma$ and $\boldsymbol E=\boldsymbol M$ gives $\sigma_{r_\ell}(\boldsymbol\Sigma+\boldsymbol M)\ge\sigma_{r_\ell}(\boldsymbol X)-\|\boldsymbol M\|_2\ge\sigma_{r_\ell}(\boldsymbol X)/2\ge\sigma_{\min}/2$, so $\boldsymbol\Sigma+\boldsymbol M$ is invertible with $\|(\boldsymbol\Sigma+\boldsymbol M)^{-1}\|_2\le2/\sigma_{\min}$. Replacing the zero block of $\boldsymbol K$ by $\boldsymbol R_U(\boldsymbol\Sigma+\boldsymbol M)^{-1}\boldsymbol R_V^\top$ gives a matrix of rank $r_\ell$, and since $\mathcal H_{r_\ell}(\boldsymbol K)$ minimizes the Frobenius distance to $\boldsymbol K$ over all matrices of rank at most $r_\ell$, we obtain
\begin{align}
    \bigl\|\boldsymbol K-\mathcal H_{r_\ell}(\boldsymbol K)\bigr\|_{\mathrm F}
    &\le\bigl\|\boldsymbol R_U(\boldsymbol\Sigma+\boldsymbol M)^{-1}\boldsymbol R_V^\top\bigr\|_{\mathrm F}
    \nonumber\\
    &\le\frac{2}{\sigma_{\min}}\|\boldsymbol R_U\|_{\mathrm F}\|\boldsymbol R_V\|_{\mathrm F}
    \le\frac{\|\boldsymbol Y\|_{\mathrm F}^2}{\sigma_{\min}},
\end{align}
which proves~\eqref{eq:retraction_error} in both regimes. For the descent inequality, let $\boldsymbol x_t^{+}=\boldsymbol x_t-\eta\widehat{\boldsymbol G}_t$. The $L$-Lipschitz gradient of $f$ and $\widehat{\boldsymbol G}_t\in T_{\boldsymbol x_t}\mathcal X$ give
\begin{equation}
    f(\boldsymbol x_t^{+})
    \le f(\boldsymbol x_t)
    -\eta\bigl\langle\operatorname{grad}f(\boldsymbol x_t),\widehat{\boldsymbol G}_t\bigr\rangle
    +\frac{L}{2}\eta^2\|\widehat{\boldsymbol G}_t\|^2.
\end{equation}
Applying~\eqref{eq:retraction_error} to each 2D block with $\boldsymbol Y=-\eta\widehat{\boldsymbol G}_{t,\ell}$ under Assumption~\ref{ass:A}(i) bounds the block retraction errors $e_\ell\ge0$. Using $\sqrt{\sum_{\ell=1}^N e_\ell^2}\le\sum_{\ell=1}^N e_\ell$ together with $\|\nabla f\|\le G_\infty$ on the retraction segment yields
\begin{equation}
    f(\boldsymbol x_{t+1})-f(\boldsymbol x_t^{+})
    \le G_\infty\|\boldsymbol x_{t+1}-\boldsymbol x_t^{+}\|
    \le\frac{2G_\infty}{\sigma_{\min}}\eta^2\|\widehat{\boldsymbol G}_t\|^2.
\end{equation}
Adding the last two inequalities gives~\eqref{eq:descent} with $L_R=L+4G_\infty/\sigma_{\min}$.
\end{proof}

\subsection*{Proof of Theorem~\ref{thm:convergence}}

\begin{proof}
We take the conditional expectation of~\eqref{eq:descent}, insert the moments of Lemma~\ref{lem:first_moment}, absorb the bias term via Young's inequality, and telescope over $t$. Taking $\mathbb E[\cdot\mid\mathcal F_t]$ in~\eqref{eq:descent} and substituting~\eqref{eq:first_moment}--\eqref{eq:second_moment} yields
\begin{align}
    \mathbb E[f(\boldsymbol x_{t+1})\mid\mathcal F_t]
    &\le
    f(\boldsymbol x_t)
    -\eta\|\operatorname{grad}f(\boldsymbol x_t)\|_{\boldsymbol D}^2
    \nonumber\\
    &\quad-\eta\langle\operatorname{grad}f(\boldsymbol x_t),\boldsymbol\beta_t\rangle
    \nonumber\\
    &\quad+L_R\eta^2(\bar D+2)\bigl(\|\operatorname{grad}f(\boldsymbol x_t)\|_{\boldsymbol D}^2+\sigma_D^2\bigr)
    \nonumber\\
    &\quad+\frac{L_R\rho_3^2}{36}\eta^2\epsilon^4(\bar D+6)^4.
\end{align}
Applying Young's inequality with~\eqref{eq:app_weighted_norm} gives $-\eta\langle\operatorname{grad}f,\boldsymbol\beta_t\rangle\le\frac{\eta}{4}\|\operatorname{grad}f\|_{\boldsymbol D}^2+\eta B_\epsilon^2$, and $\eta\le 1/(2L_R(\bar D+2))$ yields $L_R\eta^2(\bar D+2)\|\operatorname{grad}f\|_{\boldsymbol D}^2\le\frac{\eta}{2}\|\operatorname{grad}f\|_{\boldsymbol D}^2$, whence
\begin{align}
    \mathbb E[f(\boldsymbol x_{t+1})\mid\mathcal F_t]
    &\le
    f(\boldsymbol x_t)-\frac{\eta}{4}\|\operatorname{grad}f(\boldsymbol x_t)\|_{\boldsymbol D}^2
    +\eta B_\epsilon^2
    \nonumber\\
    &\quad+\eta^2L_R(\bar D+2)\sigma_D^2
    \nonumber\\
    &\quad+\eta^2\frac{L_R\rho_3^2}{36}\epsilon^4(\bar D+6)^4.
\end{align}
Taking the total expectation, telescoping over $t=t_0,\ldots,t_0+T-1$, using $f(\boldsymbol x_{t_0+T})\ge f_*$, and multiplying by $4/(\eta T)$ yields
\begin{align}
    &\frac1T\sum_{t=t_0}^{t_0+T-1}\mathbb E\bigl[\|\operatorname{grad}f(\boldsymbol x_t)\|_{\boldsymbol D}^2\bigr]
    \nonumber\\
    &\quad\le\frac{4\Delta_0}{\eta T}+4\eta L_R(\bar D+2)\sigma_D^2+4B_\epsilon^2
    \nonumber\\
    &\qquad+\eta\frac{L_R\rho_3^2}{9}\epsilon^4(\bar D+6)^4.
\end{align}
From~\eqref{eq:formal_stepsize}, we have $1/\eta\le2L_R(\bar D+2)+\sqrt{L_R(\bar D+2)\sigma_D^2T/\Delta_0}$, $\eta\sigma_D^2\le\sqrt{\Delta_0\sigma_D^2/(L_R(\bar D+2)T)}$, and $\eta\le1/(2L_R(\bar D+2))$ (including when $\sigma_D=0$). Substituting these three bounds together with $B_\epsilon=(\rho_3/6)\epsilon^2(\bar D+2)^2$ yields~\eqref{eq:formal_rate}.
\end{proof}

\bibliographystyle{IEEEtran}
\bibliography{IEEEabrv, reference}
\end{document}